\documentclass{article}

\PassOptionsToPackage{numbers,sort&compress}{natbib}

\usepackage[preprint]{neurips_2025}

\usepackage[utf8]{inputenc} %
\usepackage[T1]{fontenc}    %
\usepackage{hyperref}       %
\usepackage{url}            %
\usepackage{booktabs}       %
\usepackage{amsfonts}       %
\usepackage{nicefrac}       %
\usepackage{microtype}      %
\usepackage{xcolor}         %

\usepackage{xcolor}
\usepackage{soul}
\usepackage{bm}
\usepackage{mathtools}
\usepackage{graphicx}
\usepackage{booktabs} 
\usepackage{colortbl} 
\usepackage{array} 
\usepackage{multirow} 
\usepackage{amssymb}
\usepackage{hyperref}
\usepackage{tablefootnote}
\usepackage{float}
\usepackage{caption}
\usepackage{subfigure}

\usepackage{epstopdf}

\usepackage{algorithm}
\usepackage[noend]{algorithmic}

\usepackage{makecell}
\usepackage{array}
\usepackage{wrapfig}
\usepackage{enumitem}

\usepackage{caption}

\usepackage{amsmath}

\usepackage{hyperref} %
\usepackage{graphicx}
\usepackage{amsmath}
\usepackage{multicol}
\usepackage{xcolor}
\usepackage[most]{tcolorbox}

\usepackage{enumitem}
\usepackage{soul} 
\usepackage{color, xcolor}
\usepackage[frozencache,cachedir=minted-cache]{minted}

\definecolor{lightred}{rgb}{1, 0.8, 0.8}
\definecolor{lightgreen}{rgb}{0.8, 1, 0.8}
\definecolor{lightblue}{rgb}{0.8, 0.9, 1}
\hypersetup{
    colorlinks=true,
    linkcolor=red,
    citecolor=green,
    filecolor=magenta,
    urlcolor=cyan
}

\title{ AI Idea Bench 2025: AI Research Idea Generation Benchmark}

\author{
    Yansheng Qiu$^{1}$, Haoquan Zhang$^{2}$, Zhaopan Xu$^{4}$, Ming Li$^{2}$, Diping Song$^{2}$,\\ \textbf{ Zheng Wang$^{1}$\thanks{Equal Corresponding author.} , Kaipeng Zhang$^{2,3}$$^{\ast}$} \\
  $^{1}$School of Computer Science, Wuhan University, \\
  $^{2}$Shanghai Artificial Intelligence Laboratory \\
  $^{3}$Shanghai Innovation Institute\\
    $^{4}$ Harbin Institute of Technology
    \\
      \tt\small
  \url{https://ai-idea-bench.github.io/}
}

\begin{document}

\maketitle

\begin{abstract}
Large-scale Language Models (LLMs) have revolutionized human-AI interaction and achieved significant success in the generation of novel ideas. However, current assessments of idea generation overlook crucial factors such as knowledge leakage in LLMs, the absence of open-ended benchmarks with grounded truth, and the limited scope of feasibility analysis constrained by prompt design. These limitations hinder the potential of uncovering groundbreaking research ideas. In this paper, we present AI Idea Bench 2025, a framework designed to quantitatively evaluate and compare the ideas generated by LLMs within the domain of AI research from diverse perspectives. The framework comprises a comprehensive dataset of 3,495 AI papers and their associated inspired works, along with a robust evaluation methodology. This evaluation system gauges idea quality in two dimensions: alignment with the ground-truth content of the original papers and judgment based on general reference material. AI Idea Bench 2025’s benchmarking system stands to be an invaluable resource for assessing and comparing idea-generation techniques, thereby facilitating the automation of scientific discovery. 
\end{abstract}

\section{Introduction}
Idea generation is a fundamental pillar of scientific inquiry, propelling technological advancements and pioneering breakthroughs. Historically, this process has been largely driven by human effort, requiring expert researchers to meticulously examine a vast array of literature, identify shortcomings in current solutions, and propose novel avenues for investigation. However, the growing complexity and sheer volume of scientific literature, combined with the swift pace of technological evolution, have made this endeavor increasingly daunting for researchers. Recent progress in large language models (LLMs)~\cite{achiam2023gpt4,dubey2024llama3,yang2024qwen2,liu2024deepseek} has enabled these models to surpass human experts in various scientific disciplines, including mathematics~\cite{yu2023metamath}, theorem proving~\cite{yang2023leandojo}, and coding~\cite{chen2021evaluating}. Building on this solid scientific foundation, one might speculate that LLMs could play a pivotal role in facilitating more abstract and creative research idea generation tasks.

\begin{figure}[!t]
\centering
\includegraphics[width=0.8\textwidth]{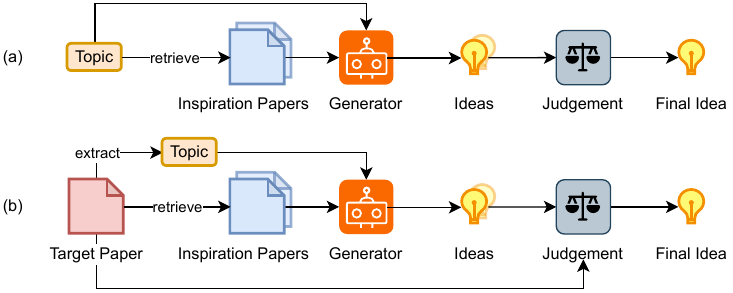}

\caption{\textbf{Comparison with current idea genearation pipline.}
(a) Current idea-generation methods retrieve relevant literature based on topics and use it as a corpus for idea generation, which leads to a lack of reference for idea evaluation. (b) Our The AI Idea Bench 2025 first identifies the target paper, then determines the corpus for idea generation by extracting its content, and uses this as the ground truth when evaluating ideas.
   }
\vspace{-6mm}
\label{fig:motivation}
\end{figure}

The exceptional performance of LLMs in various practical applications has recently garnered substantial attention in academia, particularly for their prospective role in scientific discovery and hypothesis generation~\cite{ai4science2023impact}. Numerous studies have investigated the potential of LLMs to generate hypotheses and stimulate research ideas~\cite{yang2023large2, wang2023scimon, zhou2024hypothesis, baek2024researchagent, qiu2023phenomenal}.
However, these methods encounter three major challenges in the creative generation and evaluation process: \textbf{i) There is a risk of data leakage.} Most existing approaches depend on GPT-4o as the core model for creative generation, yet they often draw upon research information that predates the latest GPT-4o training data. In scenarios where reference-free evaluation is employed, this introduces the potential for data leakage. \textbf{ii) The evaluation of creativity lacks complete ground truth.} 
Current ground-truth evaluation~\cite{guo2024ideabench} efforts are confined to the generation and assessment of titles and abstracts, omitting a thorough investigation of complete concepts—such as motivations and experiment steps, yielding incomplete reference points for evaluations that depend on large language models.
\textbf{iii) There is a deficiency in the quantitative assessment of the feasibility of ideas}, with results constrained by the limitations imposed by the design of the prompts in the judge model.

To address these challenges, we present 
AI Idea Bench 2025, a dataset and a framework designed to quantify and compare ideas generated by LLMs from multiple perspectives. Specifically, 
our dataset includes 3,495 representative papers published in AI-related conferences after October 10, 2023 \footnote{The GPT-4o, released on November 20, 2024 (gpt-4o-2024-11-20), has a knowledge cutoff of October 3, 2023.}, along with their corresponding inspiring papers. 
Additionally, we have developed an evaluation framework to assess the quality of the generated research ideas, as illustrated in Fig.~\ref{fig:motivation}. The framework is divided into two key components: The first component evaluates whether the ideas inspired by the inspiration papers align with the content of the ground truth papers. The second component involves a referenced evaluation of the idea-generation baseline. This includes comparing different baselines to rank their relative strengths and weaknesses, assessing innovation by referencing topic-related papers, and evaluating feasibility by consulting papers relevant to the experimental plans.
The benchmarking system proffered by AI Idea Bench 2025 is poised to become an invaluable resource for gauging and comparing diverse idea generation methods, ultimately propelling the automation of the scientific discovery process.
Our contributions can be summarized as follows:
\begin{itemize}
[itemsep=1pt,topsep=1pt,parsep=1pt]
\item 
We construct the AI Idea Bench 2025 dataset, comprising 3,495 influential target papers in AI-related conferences along with their corresponding motivating papers, to systematically evaluate the effectiveness of idea generation methods.
\item 
We propose an evaluation framework that aligns generated research ideas with the content of ground-truth papers, while simultaneously assessing their merits and drawbacks based on other reference material.
\item 
We conducted comprehensive experiments to showcase the effectiveness of various idea generation methods in producing innovative research ideas in AI domain, leveraging our dataset and evaluation framework.
\end{itemize}

\section{Related Work}
\subsection{Idea generation datasets}
In the past year, several studies on LLM-based scientific innovation~\cite{yang2023large2, baek2024researchagent,lu2024ai,wang2023scimon,li2024mlr,hu2024nova,wang2024scipip,su2024two,li2024chain,si2024stanfordcan} have been proposed, garnering significant attention from the LLM community.

Existing research can be categorized into four groups based on dataset construction methodologies:
i) Studies leveraging publicly available paper databases:
Wang~\emph{et al.}~\cite{wang2024scipip} established a literature database comprising 48,895 papers from major conferences including ICLR, NeurIPS, and ACL, while~\cite{su2024two} utilized the AMiner computer science dataset spanning publications from 1948 to 2014.
ii) Domain-specific literature datasets:
Wang~\emph{et al.}~\cite{wang2023scimon} developed training datasets using scientific information extracted from ACL Anthology and PubMed papers, whereas~\cite{li2024mlr,li2024chain} integrated literature through Semantic Scholar API to generate scientific hypotheses.
iii) Custom datasets via rigorous filtering:
Hu~\emph{et al.}~\cite{hu2024nova} curated a high-quality dataset of 170 papers from top-tier conferences (e.g., CVPR 2024, ACL 2024) through keyword and citation frequency filtering.
iv) Small-scale premium datasets:
Yang~\emph{et al.}~\cite{yang2023large2} constructed a dataset of 50 papers from leading social science journals, employing LLMs with triple feedback mechanisms for automated hypothesis generation.
v) Idea incomplete dataset:
Guo~\emph{et al.}~\cite{guo2024ideabench} constructed a dataset consisting of titles and abstracts from 2,374 biomedical research papers and their 29,408 reference papers, which is used to evaluate the consistency of generated ideas in terms of coherence before and after the generation process.

Many of these approaches employ GPT-4 as a tool for idea generation. However, the majority of foundational datasets lack verifiable ground truth and are predicated on open-ended evaluation paradigms. This reliance consequently introduces potential data contamination issues by disregarding the temporal limitations inherent in GPT-4's training data, thereby making it difficult to ascertain whether the generated ideas genuinely mirror current perspectives. Furthermore, some evaluation that focused exclusively on paper titles and abstracts—while neglecting the full content of the papers—inevitably leads to biased assessments.

In this paper, we collected 3,495 papers from top AI conferences published after October 3, 2023, as the ground truth, along with the main content of papers that motivated them as the input. This approach helps avoid inaccurate insight evaluation caused by potential data leakage and incomplete representation of the literature cause by limited input.

\subsection{Idea generation metric}
Most extant methodologies hinge on model-based evaluation frameworks, leveraging GPT or Claude to appraise the merit of generated ideas. Alternatively, certain approaches enlist human experts to conduct nuanced assessments.

Si~\emph{et al.}~\cite{si2024stanfordcan} introduced AI-Researcher, a system demonstrating that LLMs can conceive ideas perceived as more innovative than those crafted by human specialists. However, they caution that employing LLMs to directly appraise the multifaceted dimensions of scientific concepts yields inconsistent results.
Wang~\emph{et al.}~\cite{wang2024scipip} assessed the congruence of AI-generated ideas with empirical research by contrasting their similarity to authentic research concepts found in ACL 2024 publications, utilizing GPT-4o for evaluation. However, the scope of this Idea alignment is fixed and presents significant challenges in achieving an accurate match.
Su~\emph{et al.}~\cite{su2024two} proposed a holistic evaluation schema, quantifying the divergence of AI-produced abstracts from prior research via Historical Difference (HD). They further utilized Conformity Degree (CD) to gauge alignment with contemporary research trends, Contemporary Influence (CI) to forecast academic impact, and Overall Novelty (ON) to synthesize innovativeness and influence. Nonetheless, the brevity of thesis abstracts—often omitting nuanced rationales and detailed experimental design—limits the comprehensiveness of such analyses.

Besides, in the absence of a comparison references, the ability of LLMs to judge abstract concepts (such as good or bad, novelty, feasibility) is questionable. Moreover, when the sample size increases, the cost of human expert evaluation also rises accordingly.

In this paper, given that our proposed dataset features paired inputs alongside corresponding ground truth data, we employ the degree of concordance between the outputs generated from the input-inspired papers and the established ground truth as an objective evaluation metric. Additionally, we perform a quantitative assessment of the experimental steps involved in idea generation and determine the feasibility of the proposed ideas.
\begin{figure*}[t]
\centering
\includegraphics[width=0.95\textwidth]{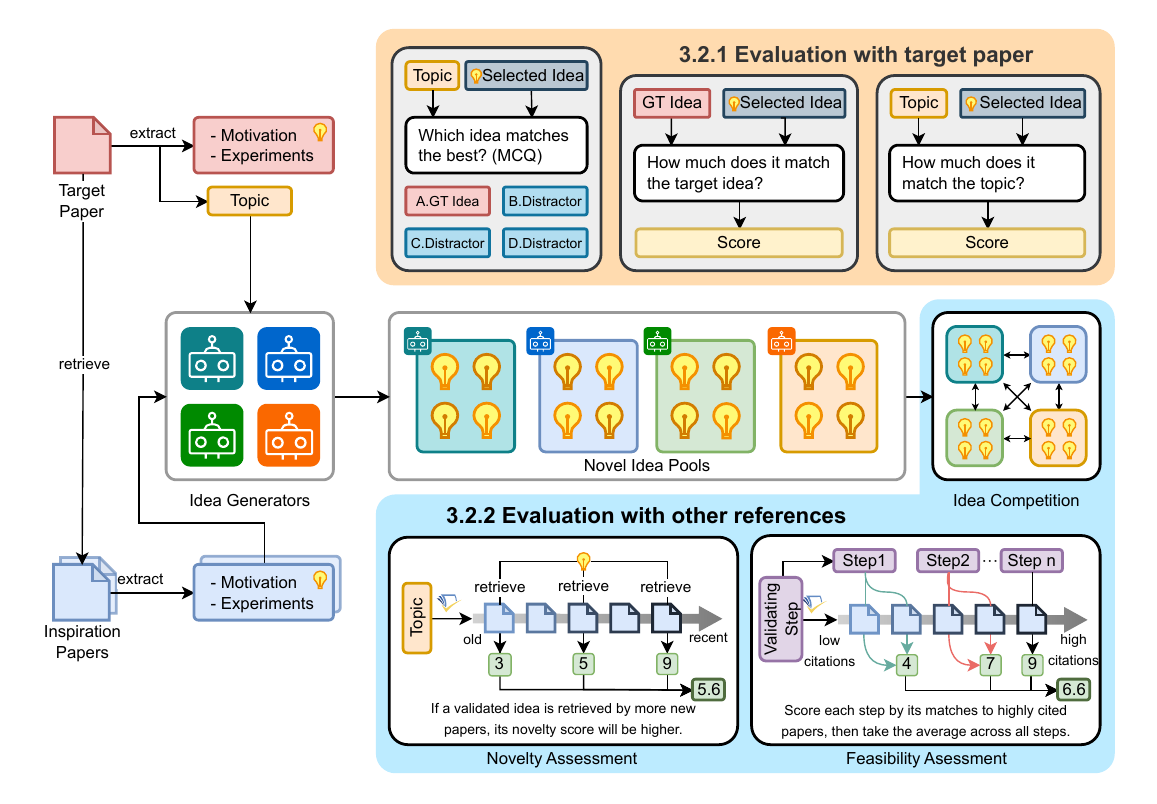}
  \caption{\textbf{Overall pipeline of AI Idea Bench 2025.} First, we decompose and summarize the motivation, experimental steps, topic, and the inspiration papers from the target paper. Then, we extract the motivation and experimental steps from the inspiration papers, and generate a cluster of ideas in combination with the topic of the target paper. Finally, we compare the idea-generation methods in six evaluations: idea multiple-choice evaluation, idea-to-idea matching, idea-to-topic matching, idea competition among baselines, novelty assessment, and feasibility assessment.}
  \vspace{-6mm}
  \label{fig:framework}
\end{figure*}
\section{AI Idea Bench 2025}
We introduce AI Idea Bench 2025, which delineates the knowledge-acquisition phase duration of base models in most idea-generation methodologies as a critical threshold. The data corresponding to publications after this period serves as the ground-truth reference. We leverage the foundational papers of these theses and user-provided background information from specific research domains to generate innovative and actionable ideas. Furthermore, we propose a comprehensive evaluation framework to assess both labeled and unlabeled baseline ideas. To achieve this, Section~\ref{sec:dataset} outlines the construction of a literature database dedicated to idea generation and data evaluation. Subsequently, in Section~\ref{sec:framework}, we provide a detailed explanation of the evaluation process using target paper and other references. The overall pipline of the AI Idea Bench 2025 is shown in Fig.~\ref{fig:framework}.

\subsection{Dataset construction}
\label{sec:dataset}
Currently, the predominant approach in human-centered research within the field of AI involves reviewing literature, identifying issues, and proposing potential solutions. This process is inherently forward-looking. However, constructing paired datasets proves challenging, as it requires careful consideration of which literature is suitable for identifying specific problems.

In this paper, we adopt an alternative research approach, seeking out papers from the literature that explicitly identify problems. Specifically, we curated papers from the top 2\% of ICLR 2025, highlights from CVPR 2024, oral presentations from ECCV 2024, spotlights and oral presentations from NeurIPS 2024, as well as spotlights and oral presentations from ICML 2024. Additionally, we included long and main presentations from NAACL 2024, EMNLP 2024, and ACL 2024. Acknowledging that some papers may release preprints on arXiv prior to formal submission, we use arXiv API~\footnote{https://info.arxiv.org/help/api/} to exclude papers published before October 3, 2023. Ultimately, we compiled a total of 3,495 papers, which will serve as the ground truth for AI Idea Bench 2025 dataset. 

As is customary, the papers that inspired  the target papers are frequently discussed across multiple sections of the text. To address this, we employed SciPDF Parser~\footnote{https://github.com/titipata/scipdf\_parser} to extract the content of the papers and utilized Deepseek V3~\cite{liu2024deepseek} to tally the cited literature. 
For each paper, We employ Deepseek V3 to extract the ten most highly cited works, then have two seasoned researchers meticulously evaluate that shortlist and select five studies deemed both feasible and reasonable as sources of literature.

Considering that the process of idea generation relies primarily on the motivation and experimental procedures of the inspiring literature, rather than its entire content, we pre-processed the inspirational papers in a manner similar to~\cite{li2024chain}. Moreover, since the outcomes of idea generation typically encompass both the motivation and experimental planning, it was necessary to extract relevant content from the target literature as well. Specifically, we summarized the problems addressed by the target literature, identifying the fields in which they were solved, as well as the methods employed to tackle these issues. During this extraction, we anonymized the methods, omitting their specific names while providing detailed descriptions of the methodological steps. Additionally, we summarized the anonymized topics of the target literature to facilitate the generation of ideas.

\subsection{Evaluation framework}
\label{sec:framework}
\subsubsection{Evaluation with target paper}
The outcomes of current idea-generation methods typically comprise motivation and experimental plans. Given that we have constructed the target papers and extracted topics, motivations, and experimental frameworks in Section~\ref{sec:dataset}, we are now in a position to objectively assess the generated results.

\textbf{Idea multiple-choice evaluation.} We first developed a multiple-choice evaluation method. Specifically,
Let \(T\) denote the target paper, with its motivation and experimental framework represented as \(M_{T}\) and \(E_{T}\), respectively. Together, these elements constitute the correct answer \(A_{c} = (M_{T}, E_{T})\) in the multiple-choice evaluation paradigm.

Consider two papers, \(L_{1}\) and \(L_{2}\), from the influential literature that exhibit the closest conceptual alignment with the target paper \(T\). Additionally, let \(L_{3}\) be the paper that maintains the highest degree of similarity to all target papers in the dataset. The set of the remaining three alternatives, denoted as \(O\), is thus given by \(O = \{L_{1}, L_{2}, L_{3}\}\). Consequently, in the multiple-choice evaluation, the complete set of answer choices is defined as \(C = \{A_{c}\} \cup O\), where \(|C| = 4\) (comprising one correct answer and three distractors).

Define \(B\) as the baseline model, which generates a cluster of ideas \(I_{B} = \{i_{1}, i_{2}, \dots, i_{n}\}\)in response to a given topic. Meanwhile, let \(F_c^D\) represent the choice function to select an option \(r_{j} \in C\), where \(D\) represent the Deepseek V3 model:
\begin{equation}
\begin{aligned}
r_{j} = {F_c^D}(i_{j}, C),
\end{aligned}
\end{equation}
All multiple-choice results of \(B\) are \(R_{B} = \{r_{1}, r_{2}, \dots, r_{n}\}\).
Given that idea generation is inherently an open-ended task, we establish a success criterion. Define the reuslt of idea multiple-choice evaluation \(S_{M}\) as a binary variable:
\begin{equation}
\begin{aligned}
S_{M}=
\begin{cases} 
1, & \text{if } A_{c} \cap R_{B}  \neq \emptyset \\ 
0, & \text{otherwise}
\end{cases},
\end{aligned}
\end{equation}
where, if \(S_{M} = 1\), we conclude that the baseline model \(B\) has successfully approximated the ground truth.

\textbf{Idea to idea matching.}
We also directly compare the generated ideas with the motivation and experimental framework of the target paper.
Specifically, define an idea similarity function
\(F_{2I}^{D}:(I_B, (M_T, E_T)) \to[0, 5]\) which measures the similarity between an idea \(i\in I_B\) and the combination of the motivation and experimental framework of the target paper \((M_T, E_T)\). For each \(i_j\in I_B\) (\(j = 1,2,\cdots,n\)), \(F_{2I}^{D}(i_j,(M_T, E_T))\) gives a score in the range \([0, 5]\).
The score \(S_{I^2}\) assigned to the current baseline \(B\) is then given by the formula:
\begin{equation}
\begin{aligned}
S_{I^2}=\max_{i_j\in I_B}F_{2I}^{D}(i_j,(M_T, E_T)).
\end{aligned}
\end{equation}

\textbf{Idea to topic matching.}
To verify that the generated ideas align with the specified topic, we also assess the degree of match between the generated ideas and the topic of the target paper.
Specifically, define a similarity function \(F_{IT}^{D}: (I_B,T_\mathrm{topic}) \to[0, 5]\), which measures the similarity between an idea \(i\in I_B\) and the topic \(T_\mathrm{topic}\) of the target paper. That is, for each \(i_j\in I_B\) (\(j = 1,2,\cdots,n\)), \(M^{IT}_{D}(i_j, T_{\mathrm{topic}})\) gives a score in the range \([0, 5]\).
The score \(S_{{IT}}\) assigned to the current baseline \(B\) for the alignment with the topic is given by the formula:
\begin{equation}
\begin{aligned}
S_{{IT}}=\max_{i_j\in I_B}F_{IT}^{D}(i_j, T_\mathrm{topic})
\end{aligned}
\end{equation}

\subsubsection{Evaluation with other references}
Given that the evaluation of abstract concepts (such as "good" or "bad") by large language models (LLMs) is inherently fraught with uncertainties in the absence of references, we incorporate references to mitigate these uncertainties when assessing the merits and demerits of ideas. This evaluation process can be categorized into two approaches: cross-referencing among baselines and cross-referencing with other literatures.

\textbf{Ideas competition among baselines.} 
For a given topic, we use Deepseek V3 to conduct pairwise comparisons of the baselines in six aspects: innovativeness, importance, quality, feasibility clarity, and topic relevance. The winner scores a point, while the loser gets no points. The total score is the sum of the scores of the current baseline.
Specifically,
Let \(B=\{b_1,b_2,\cdots,b_n\}\) ) be the set of baselines. 
Define the function \(\mathrm{Comp}(b_i,b_j)\) for \(b_i,b_j\in B\) as:
\begin{equation}
\begin{aligned}
\mathrm{Comp}(b_i,b_j)=\begin{cases}
3, & \text{if }b_i\text{ wins }b_j\\
0, & \text{if }b_i\text{ loses }b_j
\end{cases}.
\end{aligned}
\end{equation}
The total score \(S_{IC}\) of a baseline \(b_m\in B\) is given by:
\begin{equation}
\begin{aligned}
S_{IC}=\sum_{j = 1,j\neq m}^{n}\mathrm{Comp}(b_m,b_j).
\end{aligned}
\end{equation}

\textbf{Novelty assessment with reference to other literature.} Using LLMs to evaluate innovativeness is a relatively simple and straightforward approach, but it remains constrained by the knowledge cutoff inherent in LLMs. To address this, we introduce a novelty quantification method inspired by~\cite{su2024two}. Specifically, we extract keywords from the topic, establish a timeline with October 3, 2023, as the reference date and use semantic scholar API~\footnote{https://www.semanticscholar.org/product/api} to search papers. Papers published before this date are categorized as historical, while those published after are considered contemporary. Given that the information provided in paper abstracts is often incomplete and lacking specificity, we shift the focus to the ideas presented in each paper. More specifically, we compute the distance between the ideas of historical papers and the generated ideas to determine the historical difference (hd\_ideas), calculate the distance between the ideas of contemporary papers and the generated ideas to assess the contemporary difference (cd\_ideas), and use the citation count of contemporary papers to gauge their contemporary influence (cc).
Then the overall novelty socre \(S_{N}\) of an idea is: 
\begin{align}
\small
S_{N} = (1+cd\_idea)*cc/(1+hd\_idea).
\end{align} 

\textbf{Feasibility
asessment with reference to other literature.}
We continually subdivide the motivation and experimental plan of an idea until they can no longer be decomposed into two distinct concepts. The papers identified through these subdivided concepts are considered the methods that the current idea can reference. We posit that the greater the influence (measured by the number of citations) of these reference methods, the higher the feasibility of the current idea. However, when the citation count of the papers retrieved by a particular concept becomes so large that it skews the overall feasibility assessment, we introduce influence normalization to account for this:
\begin{align}
\label{eq:Influence}
Inf_N & = -e^{-\frac{x}{\lambda}}+1, 
\end{align}
where x is the number of citations which got by semantic scholar API and $\lambda$ is the impact factor, which is set to 50 in this paper.
Specifically, when the number of citations reaches 35, the influence is set at 0.5. This indicates that the paper is frequently discussed within the field and is likely used as a comparison algorithm. When the number of citations reaches 100, the influence is increased to 0.9, signifying that most innovations in the current field are likely built upon this paper.

Considering that the number of citations of a paper in the past two years reflects the influence of this paper on the current research trend, but this influence decays over time.
we propose a new influence evaluation equation:
\begin{equation}
\begin{aligned}
Inf =  
& \sum_{y_c = y_p}^{y_l-2} Inf_N(x_{y_c} )/(y_l-y_c) \\+ &\sum_{y_c = y_l-2}^{y_l} Inf_N(x_{y_c} )
\end{aligned}
\end{equation}
where $y_p$ is the published year and $y_l$ is the latest year. 
Specifically, the impact of papers from the last two years is calculated using the Equation~\ref{eq:Influence}. For papers that are further removed from the current year, their annual impact is adjusted by dividing it by the difference in years from the current year.

The feasibility \(S_F\) of an idea is determined as the average of the influences of all the papers (\(P\)) related to the concepts encompassed within the generated idea:
\begin{equation}
S_F=\frac{1}{|P|}\sum_{x\in P}Inf(x)
\end{equation}

\begin{table}[tbp]
\centering
\caption{Result of evaluation with target paper.}
\label{tb:evwtp}
\renewcommand{\arraystretch}{1}
\begin{tabular}{lcccccc}
\hline
\small
& \multicolumn{2}{c}{I2T $\uparrow$} & \multicolumn{2}{c}{I2I $\uparrow$} & \multicolumn{2}{c}{IMCQ $\uparrow$} \\
\cmidrule(lr){2-3} \cmidrule(lr){4-5} \cmidrule(lr){6-7}
& Motivation & Exp Plan & Motivation & Exp Plan & Motivation & Exp Plan \\
\hline
VIRSC & 4.974 & 4.983 & 2.937 & 2.123 & 0.509& 0.461 \\
AI-Researcher & 4.994 & 4.995 & 2.807 & 2.024 & 0.566 & 0.446 \\
AI-Scientist & 5.0 & 5.0 & 3.591 & 2.734 & 0.611 & 0.378 \\
SCIPIP & 4.986 & - & 2.437 & - & 0.595 & - \\
\hline
\end{tabular}
\vspace{-5mm}
\end{table}
\section{Experiment}
In this section, we will assess the performance of various idea-generation models using the AI Idea Bench 2025 benchmark dataset. The evaluation criteria encompass both assessments based on target papers and evaluations involving other references.
\subsection{Experimental setup}
\textbf{Dataset.}
Our dataset is built by curating high-quality papers from leading conferences. The corpus consists of 3,495 papers from CVPR 2024, ECCV 2024, ICML 2024, NeurIPS 2024, NAACL 2024, EMNLP 2024, ACL 2024, and ICLR 2025.
Additionally, we collected the inspirational source papers corresponding to the target papers, from which we constructed input-output pairs for idea generation.

\noindent\textbf{Baseline.}
To compare our proposed approach with state-of-the-art methods, we select four leading approaches as baselines: AI-Researcher~\cite{si2024stanfordcan}, AI-Scientist~\cite{lu2024ai}, 
SCIPIP~\cite{wang2024scipip}, 
and VIRSCI~\cite{su2024two}. For AI-Researcher, AI-Scientist, and SCIPIP, we run the original code on inspiration papers. Considering that SCIPIP's idea lacks the generation of a detailed experimental plan, we only use its generation results in the motivation comparison. For VIRSCI, we bypass the topic selection process and instead use the topics extracted from the target papers and the content of the motivating papers as reference inputs. To ensure consistency and avoid discrepancies in idea generation results, we use GPT-4o-2024-11-20 as the base model for all methods, unifying the knowledge cutoff date of the model. For each baseline we generate two ideas in the same inspiration papers.

\subsection{Main results}
\label{main_result}
\subsubsection{Evaluation with target paper}

The primary objective of this sub-evaluation framework is to ascertain whether extant idea-generation models can produce outputs that align conceptually with target papers when provided with inspirational papers.
The Idea Multiple-Choice Evaluation (IMCQ) and Idea-to-Idea Matching (I2I) methodologies are designed to assess the degree of congruence between generated ideas and the motivation and experiment steps outlined in the target papers. 
The distinction between these two methods lies in the nature of the comparison: in IMCQ, some answer choices are partially drawn from inspirational literature, enabling us to evaluate whether the generation method can transcend the limitations of its input.
Conversely, I2I assesses the resemblance between generated and target ideas through a more nuanced lens, evaluating:
i) For motivation: Core Issue Analysis, Contextual Alignment, and Structural/Theoretical Overlap.
ii)For experimental design: Structural Similarity, Theoretical Consistency, and Problem-Centric Focus.
The Idea-to-Topic Matching (I2T) protocol, conversely, evaluates whether the motivational and experimental components of generated ideas remain consistently aligned with the specified input topic throughout the generative process.

As shown in Table~\ref{tb:evwtp}, ideas generated by AI-Scientist demonstrate the highest degree of alignment with those of the target paper. In terms of departing from the constraints of the inspirational literature,  AI-Scientist exhibit the strongest insight ability in motivation, while VIRSC perform better in Experimental plans. 
With regard to topic relevance, all baseline models display an exceptionally high level of consistency with the thematic focus of the target paper.

\begin{table}[tbp]
\centering
\renewcommand{\arraystretch}{1}
\caption{Result of evaluation with other references.}
\label{tb:evor}
\resizebox{\textwidth}{!}{
\begin{tabular}{lcccccc}
\toprule
 & \multicolumn{2}{c}{{IC}} & \multicolumn{2}{c}{{NA $\uparrow$}}  & {FA $\uparrow$} & {FPS $\uparrow$}\\
\cmidrule(lr){2-3} \cmidrule(lr){4-5}
 &Total Rank $\downarrow$ & Total Score $\uparrow$ & Motivation & Exp Plan &  &\\
\midrule
VIRSC  & 7211 & 23787 & 24.873 & 24.654  & 0.133 &$8.290\times10^{-3}$\\
AI-Researcher & 4475 & 29345 & 24.917 & 24.692  & 0.168 &$9.728\times10^{-3}$\\
    AI-Scientist & 9537 & 19195 & 25.030 & 26.080  & 0.121 &$17.003\times10^{-3}$\\
SCIPIP & 13362 & 11553 & 25.055 & --  & -- &--\\
\bottomrule
\end{tabular}
}
\vspace{-5mm}
\end{table}
\vspace{-3mm}
\subsubsection{Evaluation with other references}

The primary purpose of this evaluation procedure is to assess the capabilities of the ideas generated by baselines beyond the scope of the target paper.
The Ideas Competition (IC) treats the outputs of each baseline as mutual points of reference, aggregating their rankings and scores to identify the most effective method among them. The Novelty Assessment (NA) measures the similarity between the ideas produced by the baselines and those found in both historical and contemporary literature, thereby evaluating the originality of the generated content.
The Feasibility Assessment (FA) focuses on determining whether the proposed experimental approaches are grounded in established, effective methodologies. Complementarily, Feasibility Per Step (FPS) evaluates the methodological soundness of each individual step within the proposed experimental framework.

As presented in Table~\ref{tb:evor}, AI-Researcher significantly outperforms the other baselines in both the circular comparison and overall feasibility assessment. AI-Scientist, on the other hand, demonstrates superior performance in terms of novelty and exhibits greater average feasibility at the step level within the experimental design.

\begin{figure*}[th!]
\centering
\includegraphics[width=0.95\textwidth]{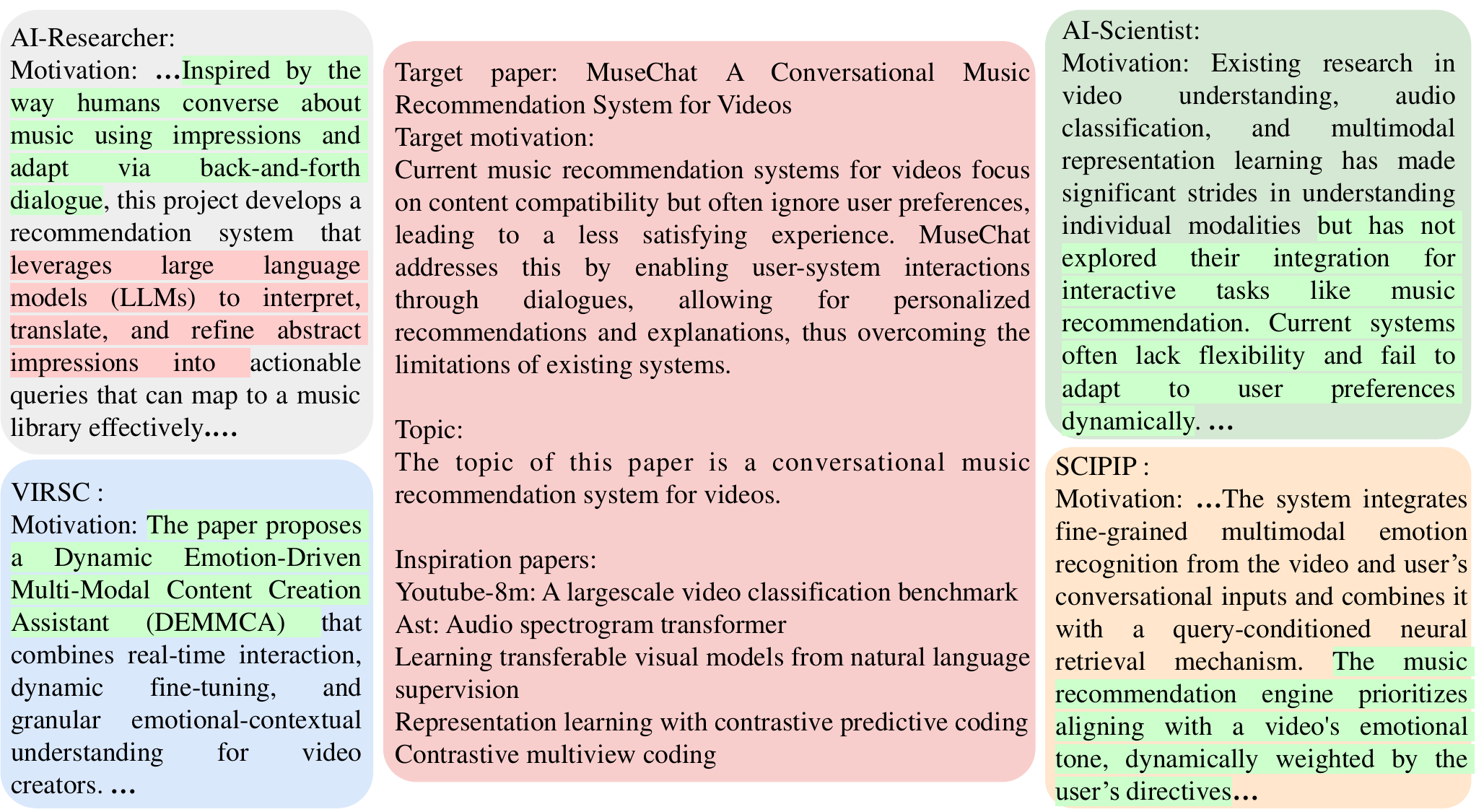}
  \caption{\textbf{A case of idea generation on motivation.} In the visual annotations, text highlighted with a \sethlcolor{lightgreen}\hl{green} background denotes areas of overlap between the generated ideas and those of the target paper. The \sethlcolor{lightred}\hl{red} background indicates elements within the generated ideas that are thematically aligned with current research based on the given topic.} 
  \vspace{-3mm}
  \label{fig:case}
\end{figure*}
\vspace{-3mm}
\subsection{Case study and analysis}
In Fig.~\ref{fig:case}, we show an example of generating ideas, mainly showing the motivation part. And we also present some case studies, each comprising the content of the target paper, the inspirational papers that informed idea generation, and the corresponding outputs produced by each baseline method in Appendix~\ref{appendix:Case_0}–\ref{appendix:Case_4}.

In most instances, AI-Scientist exhibits a greater degree of overlap with the target ideas. This can be attributed to its use of large language models to score and re-rank all input papers based on their direct relevance to the specified topic, thereby achieving closer alignment with the core ideas of the target papers.
Following closely are VIRSC and AI-Researcher. VIRSC benefits from a multi-round, multi-agent discussion mechanism, which helps mitigate bias during idea generation. In contrast, AI-Researcher preserves previously generated ideas and employs multi-stage reasoning chains combined with self-reflection to iteratively refine and expand upon each idea, thereby enhancing coherence and conceptual stability.
A notable limitation of SCIPIP lies in its inability to deconstruct the components of an idea, resulting in outputs that lack a coherent structural framework and are difficult to interpret from their textual descriptions.
Furthermore, we observed that AI-Scientist and AI-Researcher tend to incorporate specific algorithmic choices and detailed comparative strategies within their experimental designs. Although these additions did not yield significant advantages in our evaluation metrics, they may offer researchers more granular and actionable guidance in experimental planning.

While SCIPIP, AI-Scientist, and AI-Researcher all employ Semantic Scholar’s API for supplementary evaluation in idea generation, their methods of integration diverge significantly. AI-Scientist enriches the generation prompt by embedding selected supplementary literature, thereby providing the model with a more informed context. In contrast, AI-Researcher utilizes the retrieved literature solely as an auxiliary resource for evaluating novelty during the generation process, without incorporating it into the prompt itself. SCIPIP, meanwhile, reconstructs an entirely new research background by synthesizing information from both the input and the retrieved literature. We believe these differing approaches account for the notable disparities in their performance.
In the case of VIRSC, its dependence on a locally stored database—devoid of real-time updates—imposes a significant constraint on the novelty of its outputs, limiting the method’s capacity to generate ideas that reflect the latest developments in the field.

Based on the above analysis, we summarize the key factors that currently underpin the generation of useful and impactful research ideas:
\vspace{-2mm}
\begin{itemize}
    \item Accurate inspirational papers need to be retrieved, which can be done manually or by retrieving a large number of papers and using an LLM for relevance scoring. 
    \item Ideas should be generated in multiple rounds, with each generated idea archived and used as input for subsequent idea generation.
    \item Evaluate the generated ideas by retrieving papers based on both the ideas and the topic.
    \item Clarify the framework and components of the idea to reduce the difficulty of generation.
\end{itemize}
\vspace{-5mm}
\section{Conclusion}

In this paper, we present AI Idea Bench 2025, a comprehensive benchmark dataset and evaluation framework designed to assess the ability of existing idea generation methods to produce ideas that align with those of target papers when given inspirational sources, while also evaluating the novelty and feasibility of the generated content.
The AI Idea Bench 2025 dataset comprises 3,495 papers from premier AI conferences, all published after the knowledge cutoff date of the underlying generation model, thereby eliminating the risk of knowledge leakage.
The accompanying evaluation framework operates through a reference-based assessment paradigm—drawing on the target paper’s ideas, ideas generated by alternative methods, and the broader body of published literature—while offering an interpretable, open-ended evaluation pipeline.
This work aims to furnish the research community with robust, quantitative tools for conducting idea discovery research powered by large language models.

\label{sec:conclusion}

\small
\bibliography{custom}

\begin{thebibliography}{50}
\providecommand{\natexlab}[1]{#1}
\providecommand{\url}[1]{\texttt{#1}}
\expandafter\ifx\csname urlstyle\endcsname\relax
  \providecommand{\doi}[1]{doi: #1}\else
  \providecommand{\doi}{doi: \begingroup \urlstyle{rm}\Url}\fi

\bibitem[Achiam et~al.(2023)Achiam, Adler, Agarwal, Ahmad, Akkaya, Aleman, Almeida, Altenschmidt, Altman, Anadkat, et~al.]{achiam2023gpt4}
Josh Achiam, Steven Adler, Sandhini Agarwal, Lama Ahmad, Ilge Akkaya, Florencia~Leoni Aleman, Diogo Almeida, Janko Altenschmidt, Sam Altman, Shyamal Anadkat, et~al.
\newblock Gpt-4 technical report.
\newblock \emph{arXiv preprint arXiv:2303.08774}, 2023.

\bibitem[Dubey et~al.(2024)Dubey, Jauhri, Pandey, Kadian, Al-Dahle, Letman, Mathur, Schelten, Yang, Fan, et~al.]{dubey2024llama3}
Abhimanyu Dubey, Abhinav Jauhri, Abhinav Pandey, Abhishek Kadian, Ahmad Al-Dahle, Aiesha Letman, Akhil Mathur, Alan Schelten, Amy Yang, Angela Fan, et~al.
\newblock The llama 3 herd of models.
\newblock \emph{arXiv preprint arXiv:2407.21783}, 2024.

\bibitem[Yang et~al.(2024)Yang, Yang, Hui, Zheng, Yu, Zhou, Li, Li, Liu, Huang, et~al.]{yang2024qwen2}
An~Yang, Baosong Yang, Binyuan Hui, Bo~Zheng, Bowen Yu, Chang Zhou, Chengpeng Li, Chengyuan Li, Dayiheng Liu, Fei Huang, et~al.
\newblock Qwen2 technical report.
\newblock \emph{arXiv preprint arXiv:2407.10671}, 2024.

\bibitem[Liu et~al.(2024)Liu, Feng, Xue, Wang, Wu, Lu, Zhao, Deng, Zhang, Ruan, et~al.]{liu2024deepseek}
Aixin Liu, Bei Feng, Bing Xue, Bingxuan Wang, Bochao Wu, Chengda Lu, Chenggang Zhao, Chengqi Deng, Chenyu Zhang, Chong Ruan, et~al.
\newblock Deepseek-v3 technical report.
\newblock \emph{arXiv preprint arXiv:2412.19437}, 2024.

\bibitem[Yu et~al.(2023)Yu, Jiang, Shi, Yu, Liu, Zhang, Kwok, Li, Weller, and Liu]{yu2023metamath}
Longhui Yu, Weisen Jiang, Han Shi, Jincheng Yu, Zhengying Liu, Yu~Zhang, James~T Kwok, Zhenguo Li, Adrian Weller, and Weiyang Liu.
\newblock Metamath: Bootstrap your own mathematical questions for large language models.
\newblock \emph{arXiv preprint arXiv:2309.12284}, 2023.

\bibitem[Yang et~al.(2023{\natexlab{a}})Yang, Swope, Gu, Chalamala, Song, Yu, Godil, Prenger, and Anandkumar]{yang2023leandojo}
Kaiyu Yang, Aidan~M Swope, Alex Gu, Rahul Chalamala, Peiyang Song, Shixing Yu, Saad Godil, Ryan Prenger, and Anima Anandkumar.
\newblock Leandojo: Theorem proving with retrieval-augmented language models.
\newblock In \emph{Thirty-seventh Conference on Neural Information Processing Systems Datasets and Benchmarks Track}, 2023{\natexlab{a}}.
\newblock URL \url{https://openreview.net/forum?id=g7OX2sOJtn}.

\bibitem[Chen et~al.(2021)Chen, Tworek, Jun, Yuan, Pinto, Kaplan, Edwards, Burda, Joseph, Brockman, et~al.]{chen2021evaluating}
Mark Chen, Jerry Tworek, Heewoo Jun, Qiming Yuan, Henrique Ponde De~Oliveira Pinto, Jared Kaplan, Harri Edwards, Yuri Burda, Nicholas Joseph, Greg Brockman, et~al.
\newblock Evaluating large language models trained on code.
\newblock \emph{arXiv preprint arXiv:2107.03374}, 2021.

\bibitem[AI4Science and Quantum(2023)]{ai4science2023impact}
Microsoft~Research AI4Science and Microsoft~Azure Quantum.
\newblock The impact of large language models on scientific discovery: a preliminary study using gpt-4.
\newblock \emph{arXiv preprint arXiv:2311.07361}, 2023.

\bibitem[Yang et~al.(2023{\natexlab{b}})Yang, Du, Li, Zheng, Poria, and Cambria]{yang2023large2}
Zonglin Yang, Xinya Du, Junxian Li, Jie Zheng, Soujanya Poria, and Erik Cambria.
\newblock Large language models for automated open-domain scientific hypotheses discovery.
\newblock \emph{arXiv preprint arXiv:2309.02726}, 2023{\natexlab{b}}.

\bibitem[Wang et~al.(2023{\natexlab{a}})Wang, Downey, Ji, and Hope]{wang2023scimon}
Qingyun Wang, Doug Downey, Heng Ji, and Tom Hope.
\newblock Scimon: Scientific inspiration machines optimized for novelty.
\newblock \emph{arXiv preprint arXiv:2305.14259}, 2023{\natexlab{a}}.

\bibitem[Zhou et~al.(2024)Zhou, Liu, Srivastava, Mei, and Tan]{zhou2024hypothesis}
Yangqiaoyu Zhou, Haokun Liu, Tejes Srivastava, Hongyuan Mei, and Chenhao Tan.
\newblock Hypothesis generation with large language models.
\newblock \emph{arXiv preprint arXiv:2404.04326}, 2024.

\bibitem[Baek et~al.(2024)Baek, Jauhar, Cucerzan, and Hwang]{baek2024researchagent}
Jinheon Baek, Sujay~Kumar Jauhar, Silviu Cucerzan, and Sung~Ju Hwang.
\newblock Researchagent: Iterative research idea generation over scientific literature with large language models.
\newblock \emph{arXiv preprint arXiv:2404.07738}, 2024.

\bibitem[Qiu et~al.(2023)Qiu, Jiang, Lu, Sclar, Pyatkin, Bhagavatula, Wang, Kim, Choi, Dziri, et~al.]{qiu2023phenomenal}
Linlu Qiu, Liwei Jiang, Ximing Lu, Melanie Sclar, Valentina Pyatkin, Chandra Bhagavatula, Bailin Wang, Yoon Kim, Yejin Choi, Nouha Dziri, et~al.
\newblock Phenomenal yet puzzling: Testing inductive reasoning capabilities of language models with hypothesis refinement.
\newblock \emph{arXiv preprint arXiv:2310.08559}, 2023.

\bibitem[Guo et~al.(2024)Guo, Shariatmadari, Xiong, Huang, Xie, Bekiranov, and Zhang]{guo2024ideabench}
Sikun Guo, Amir~Hassan Shariatmadari, Guangzhi Xiong, Albert Huang, Eric Xie, Stefan Bekiranov, and Aidong Zhang.
\newblock Ideabench: Benchmarking large language models for research idea generation.
\newblock \emph{arXiv preprint arXiv:2411.02429}, 2024.

\bibitem[Lu et~al.(2024)Lu, Lu, Lange, Foerster, Clune, and Ha]{lu2024ai}
Chris Lu, Cong Lu, Robert~Tjarko Lange, Jakob Foerster, Jeff Clune, and David Ha.
\newblock The ai scientist: Towards fully automated open-ended scientific discovery.
\newblock \emph{arXiv preprint arXiv:2408.06292}, 2024.

\bibitem[Li et~al.(2024{\natexlab{a}})Li, Patel, Wang, and Du]{li2024mlr}
Ruochen Li, Teerth Patel, Qingyun Wang, and Xinya Du.
\newblock Mlr-copilot: Autonomous machine learning research based on large language models agents.
\newblock \emph{arXiv preprint arXiv:2408.14033}, 2024{\natexlab{a}}.

\bibitem[Hu et~al.(2024)Hu, Fu, Wang, Wang, Li, Xu, Lu, Jin, Pan, and Lan]{hu2024nova}
Xiang Hu, Hongyu Fu, Jinge Wang, Yifeng Wang, Zhikun Li, Renjun Xu, Yu~Lu, Yaochu Jin, Lili Pan, and Zhenzhong Lan.
\newblock Nova: An iterative planning and search approach to enhance novelty and diversity of llm generated ideas.
\newblock \emph{arXiv preprint arXiv:2410.14255}, 2024.

\bibitem[Wang et~al.(2024)Wang, Gu, Zhang, Luo, Dai, Shen, Xie, Lin, He, and Ye]{wang2024scipip}
Wenxiao Wang, Lihui Gu, Liye Zhang, Yunxiang Luo, Yi~Dai, Chen Shen, Liang Xie, Binbin Lin, Xiaofei He, and Jieping Ye.
\newblock Scipip: An llm-based scientific paper idea proposer.
\newblock \emph{arXiv preprint arXiv:2410.23166}, 2024.

\bibitem[Su et~al.(2024)Su, Chen, Tang, Zheng, Li, Yin, Ouyang, and Dong]{su2024two}
Haoyang Su, Renqi Chen, Shixiang Tang, Xinzhe Zheng, Jingzhe Li, Zhenfei Yin, Wanli Ouyang, and Nanqing Dong.
\newblock Two heads are better than one: A multi-agent system has the potential to improve scientific idea generation.
\newblock \emph{arXiv preprint arXiv:2410.09403}, 2024.

\bibitem[Li et~al.(2024{\natexlab{b}})Li, Xu, Guo, Zhao, Li, Yuan, Zhang, Jiang, Xin, Dang, et~al.]{li2024chain}
Long Li, Weiwen Xu, Jiayan Guo, Ruochen Zhao, Xingxuan Li, Yuqian Yuan, Boqiang Zhang, Yuming Jiang, Yifei Xin, Ronghao Dang, et~al.
\newblock Chain of ideas: Revolutionizing research via novel idea development with llm agents.
\newblock \emph{arXiv preprint arXiv:2410.13185}, 2024{\natexlab{b}}.

\bibitem[Si et~al.(2024)Si, Yang, and Hashimoto]{si2024stanfordcan}
Chenglei Si, Diyi Yang, and Tatsunori Hashimoto.
\newblock Can llms generate novel research ideas? a large-scale human study with 100+ nlp researchers.
\newblock \emph{arXiv preprint arXiv:2409.04109}, 2024.

\bibitem[Kong et~al.(2024)Kong, Liu, Ng, Cottereau, and Ooi]{kong2024openess}
Lingdong Kong, Youquan Liu, Lai~Xing Ng, Benoit~R Cottereau, and Wei~Tsang Ooi.
\newblock Openess: Event-based semantic scene understanding with open vocabularies.
\newblock In \emph{Proceedings of the IEEE/CVF Conference on Computer Vision and Pattern Recognition}, pages 15686--15698, 2024.

\bibitem[Radford et~al.(2021)Radford, Kim, Hallacy, Ramesh, Goh, Agarwal, Sastry, Askell, Mishkin, Clark, et~al.]{radford2021learning}
Alec Radford, Jong~Wook Kim, Chris Hallacy, Aditya Ramesh, Gabriel Goh, Sandhini Agarwal, Girish Sastry, Amanda Askell, Pamela Mishkin, Jack Clark, et~al.
\newblock Learning transferable visual models from natural language supervision.
\newblock In \emph{International conference on machine learning}, pages 8748--8763. PmLR, 2021.

\bibitem[Alonso and Murillo(2019)]{alonso2019ev}
Inigo Alonso and Ana~C Murillo.
\newblock Ev-segnet: Semantic segmentation for event-based cameras.
\newblock In \emph{Proceedings of the IEEE/CVF Conference on Computer Vision and Pattern Recognition Workshops}, pages 0--0, 2019.

\bibitem[Gallego et~al.(2020)Gallego, Delbr{\"u}ck, Orchard, Bartolozzi, Taba, Censi, Leutenegger, Davison, Conradt, Daniilidis, et~al.]{gallego2020event}
Guillermo Gallego, Tobi Delbr{\"u}ck, Garrick Orchard, Chiara Bartolozzi, Brian Taba, Andrea Censi, Stefan Leutenegger, Andrew~J Davison, J{\"o}rg Conradt, Kostas Daniilidis, et~al.
\newblock Event-based vision: A survey.
\newblock \emph{IEEE transactions on pattern analysis and machine intelligence}, 44\penalty0 (1):\penalty0 154--180, 2020.

\bibitem[Sun et~al.(2022)Sun, Messikommer, Gehrig, and Scaramuzza]{sun2022ess}
Zhaoning Sun, Nico Messikommer, Daniel Gehrig, and Davide Scaramuzza.
\newblock Ess: Learning event-based semantic segmentation from still images.
\newblock In \emph{European Conference on Computer Vision}, pages 341--357. Springer, 2022.

\bibitem[Achanta et~al.(2012)Achanta, Shaji, Smith, Lucchi, Fua, and S{\"u}sstrunk]{achanta2012slic}
Radhakrishna Achanta, Appu Shaji, Kevin Smith, Aurelien Lucchi, Pascal Fua, and Sabine S{\"u}sstrunk.
\newblock Slic superpixels compared to state-of-the-art superpixel methods.
\newblock \emph{IEEE transactions on pattern analysis and machine intelligence}, 34\penalty0 (11):\penalty0 2274--2282, 2012.

\bibitem[Yang et~al.()Yang, Yu, Zhang, Cao, Xu, Zhang, Gonzalez, and Cui]{yang2406buffer}
L~Yang, Z~Yu, T~Zhang, S~Cao, M~Xu, W~Zhang, JE~Gonzalez, and B~Cui.
\newblock Buffer of thoughts: Thought-augmented reasoning with large language models, 2024.
\newblock \emph{URL https://arxiv. org/abs/2406.04271}.

\bibitem[Wei et~al.(2022)Wei, Wang, Schuurmans, Bosma, Xia, Chi, Le, Zhou, et~al.]{wei2022chain}
Jason Wei, Xuezhi Wang, Dale Schuurmans, Maarten Bosma, Fei Xia, Ed~Chi, Quoc~V Le, Denny Zhou, et~al.
\newblock Chain-of-thought prompting elicits reasoning in large language models.
\newblock \emph{Advances in neural information processing systems}, 35:\penalty0 24824--24837, 2022.

\bibitem[Yao et~al.(2023)Yao, Yu, Zhao, Shafran, Griffiths, Cao, and Narasimhan]{yao2023tree}
Shunyu Yao, Dian Yu, Jeffrey Zhao, Izhak Shafran, Tom Griffiths, Yuan Cao, and Karthik Narasimhan.
\newblock Tree of thoughts: Deliberate problem solving with large language models.
\newblock \emph{Advances in neural information processing systems}, 36:\penalty0 11809--11822, 2023.

\bibitem[Xu et~al.(2023)Xu, Yang, Lin, Wang, Zhou, Zhang, and Mao]{xu2023expertprompting}
Benfeng Xu, An~Yang, Junyang Lin, Quan Wang, Chang Zhou, Yongdong Zhang, and Zhendong Mao.
\newblock Expertprompting: Instructing large language models to be distinguished experts.
\newblock \emph{arXiv preprint arXiv:2305.14688}, 2023.

\bibitem[Gao et~al.(2023)Gao, Madaan, Zhou, Alon, Liu, Yang, Callan, and Neubig]{gao2023pal}
Luyu Gao, Aman Madaan, Shuyan Zhou, Uri Alon, Pengfei Liu, Yiming Yang, Jamie Callan, and Graham Neubig.
\newblock Pal: Program-aided language models.
\newblock In \emph{International Conference on Machine Learning}, pages 10764--10799. PMLR, 2023.

\bibitem[Wang et~al.(2022)Wang, Wei, Schuurmans, Le, Chi, Narang, Chowdhery, and Zhou]{wang2022self}
Xuezhi Wang, Jason Wei, Dale Schuurmans, Quoc Le, Ed~Chi, Sharan Narang, Aakanksha Chowdhery, and Denny Zhou.
\newblock Self-consistency improves chain of thought reasoning in language models.
\newblock \emph{arXiv preprint arXiv:2203.11171}, 2022.

\bibitem[Agrawal et~al.(2022)Agrawal, Hegselmann, Lang, Kim, and Sontag]{agrawal2022large}
Monica Agrawal, Stefan Hegselmann, Hunter Lang, Yoon Kim, and David Sontag.
\newblock Large language models are few-shot clinical information extractors.
\newblock In \emph{Proceedings of the 2022 Conference on Empirical Methods in Natural Language Processing}, pages 1998--2022, 2022.

\bibitem[Al-Ars et~al.(2023)Al-Ars, Agba, Guo, Boerkamp, Jaber, and Jaber]{al2023nlice}
Zaid Al-Ars, Obinna Agba, Zhuoran Guo, Christiaan Boerkamp, Ziyaad Jaber, and Tareq Jaber.
\newblock Nlice: Synthetic medical record generation for effective primary healthcare differential diagnosis.
\newblock In \emph{2023 IEEE 23rd International Conference on Bioinformatics and Bioengineering (BIBE)}, pages 397--402. IEEE, 2023.

\bibitem[Bae et~al.(2023)Bae, Kyung, Ryu, Cho, Lee, Kweon, Oh, Ji, Chang, Kim, et~al.]{bae2023ehrxqa}
Seongsu Bae, Daeun Kyung, Jaehee Ryu, Eunbyeol Cho, Gyubok Lee, Sunjun Kweon, Jungwoo Oh, Lei Ji, Eric Chang, Tackeun Kim, et~al.
\newblock Ehrxqa: A multi-modal question answering dataset for electronic health records with chest x-ray images.
\newblock \emph{Advances in Neural Information Processing Systems}, 36:\penalty0 3867--3880, 2023.

\bibitem[Barnett et~al.(2019)Barnett, Boddupalli, Nundy, and Bates]{barnett2019comparative}
Michael~L Barnett, Dhruv Boddupalli, Shantanu Nundy, and David~W Bates.
\newblock Comparative accuracy of diagnosis by collective intelligence of multiple physicians vs individual physicians.
\newblock \emph{JAMA network open}, 2\penalty0 (3):\penalty0 e190096--e190096, 2019.

\bibitem[Ben-Assuli et~al.(2020)Ben-Assuli, Kumar, Arazy, and Shabtai]{ben2020use}
Ofir Ben-Assuli, Nanda Kumar, Ofer Arazy, and Itamar Shabtai.
\newblock The use of analytic hierarchy process for measuring the complexity of medical diagnosis.
\newblock \emph{Health Informatics Journal}, 26\penalty0 (1):\penalty0 218--232, 2020.

\bibitem[Tan et~al.(2024)Tan, Sun, Hu, Wang, Deilamsalehy, Plummer, Russell, and Saenko]{tan2024koala}
Reuben Tan, Ximeng Sun, Ping Hu, Jui-hsien Wang, Hanieh Deilamsalehy, Bryan~A Plummer, Bryan Russell, and Kate Saenko.
\newblock Koala: Key frame-conditioned long video-llm.
\newblock In \emph{Proceedings of the IEEE/CVF Conference on Computer Vision and Pattern Recognition}, pages 13581--13591, 2024.

\bibitem[Li et~al.(2023)Li, He, Wang, Li, Wang, Luo, Wang, Wang, and Qiao]{li2023videochat}
KunChang Li, Yinan He, Yi~Wang, Yizhuo Li, Wenhai Wang, Ping Luo, Yali Wang, Limin Wang, and Yu~Qiao.
\newblock Videochat: Chat-centric video understanding.
\newblock \emph{arXiv preprint arXiv:2305.06355}, 2023.

\bibitem[Mangalam et~al.(2023)Mangalam, Akshulakov, and Malik]{mangalam2023egoschema}
Karttikeya Mangalam, Raiymbek Akshulakov, and Jitendra Malik.
\newblock Egoschema: A diagnostic benchmark for very long-form video language understanding.
\newblock \emph{Advances in Neural Information Processing Systems}, 36:\penalty0 46212--46244, 2023.

\bibitem[Miech et~al.(2019)Miech, Zhukov, Alayrac, Tapaswi, Laptev, and Sivic]{miech2019howto100m}
Antoine Miech, Dimitri Zhukov, Jean-Baptiste Alayrac, Makarand Tapaswi, Ivan Laptev, and Josef Sivic.
\newblock Howto100m: Learning a text-video embedding by watching hundred million narrated video clips.
\newblock In \emph{Proceedings of the IEEE/CVF international conference on computer vision}, pages 2630--2640, 2019.

\bibitem[Zhang et~al.(2023)Zhang, Li, and Bing]{zhang2023video}
Hang Zhang, Xin Li, and Lidong Bing.
\newblock Video-llama: An instruction-tuned audio-visual language model for video understanding.
\newblock \emph{arXiv preprint arXiv:2306.02858}, 2023.

\bibitem[Wang et~al.(2023{\natexlab{b}})Wang, Xu, Cui, Wan, and Zhang]{wang2023fine}
Xiaolong Wang, Runsen Xu, Zhuofan Cui, Zeyu Wan, and Yu~Zhang.
\newblock Fine-grained cross-view geo-localization using a correlation-aware homography estimator.
\newblock \emph{Advances in Neural Information Processing Systems}, 36:\penalty0 5301--5319, 2023{\natexlab{b}}.

\bibitem[Zhang et~al.(2024)Zhang, Diao, Lin, Fung, Lian, Wang, Chen, Ji, and Zhang]{zhang2024r}
Hanning Zhang, Shizhe Diao, Yong Lin, Yi~Fung, Qing Lian, Xingyao Wang, Yangyi Chen, Heng Ji, and Tong Zhang.
\newblock R-tuning: Instructing large language models to say ‘i don’t know’.
\newblock In \emph{Proceedings of the 2024 Conference of the North American Chapter of the Association for Computational Linguistics: Human Language Technologies (Volume 1: Long Papers)}, pages 7106--7132, 2024.

\bibitem[Manakul et~al.(2023)Manakul, Liusie, and Gales]{manakul2023selfcheckgpt}
Potsawee Manakul, Adian Liusie, and Mark Gales.
\newblock Selfcheckgpt: Zero-resource black-box hallucination detection for generative large language models.
\newblock In \emph{Proceedings of the 2023 Conference on Empirical Methods in Natural Language Processing}, pages 9004--9017, 2023.

\bibitem[Maynez et~al.(2020)Maynez, Narayan, Bohnet, and McDonald]{maynez2020faithfulness}
Joshua Maynez, Shashi Narayan, Bernd Bohnet, and Ryan McDonald.
\newblock On faithfulness and factuality in abstractive summarization.
\newblock \emph{arXiv preprint arXiv:2005.00661}, 2020.

\bibitem[Hendrycks et~al.(2020)Hendrycks, Burns, Basart, Zou, Mazeika, Song, and Steinhardt]{hendrycks2020measuring}
Dan Hendrycks, Collin Burns, Steven Basart, Andy Zou, Mantas Mazeika, Dawn Song, and Jacob Steinhardt.
\newblock Measuring massive multitask language understanding.
\newblock \emph{arXiv preprint arXiv:2009.03300}, 2020.

\bibitem[Guo et~al.(2017)Guo, Pleiss, Sun, and Weinberger]{guo2017calibration}
Chuan Guo, Geoff Pleiss, Yu~Sun, and Kilian~Q Weinberger.
\newblock On calibration of modern neural networks.
\newblock In \emph{International conference on machine learning}, pages 1321--1330. PMLR, 2017.

\bibitem[Touvron et~al.(2023)Touvron, Lavril, Izacard, Martinet, Lachaux, Lacroix, Rozi{\`e}re, Goyal, Hambro, Azhar, et~al.]{touvron2023llama}
Hugo Touvron, Thibaut Lavril, Gautier Izacard, Xavier Martinet, Marie-Anne Lachaux, Timoth{\'e}e Lacroix, Baptiste Rozi{\`e}re, Naman Goyal, Eric Hambro, Faisal Azhar, et~al.
\newblock Llama: Open and efficient foundation language models.
\newblock \emph{arXiv preprint arXiv:2302.13971}, 2023.

\end{thebibliography}

\newpage

\normalsize
\appendix

\setcounter{theorem}{0} 
\setcounter{lemma}{0}   
\setcounter{proposition}{0}

\tableofcontents

\newpage

\onecolumn

\section{Limitation}
\label{sec:limitation}
This paper introduces AI Idea Bench 2025, which aims to evaluate the ability of existing creative generation methods to generate ideas consistent with target papers when given an inspiration source. While this represents a crucial step for AI in science, there are still other unexplored parts of the entire research pipeline. In our subsequent work, we will conduct research on the deployment and implementation of these ideas.

\section{Case 0}
\label{appendix:Case_0}
\begin{tcolorbox}[breakable,enhanced,width=\textwidth]
\textbf{Target paper:} 

Openess: Event-based semantic scene understanding with open vocabularies~\cite{kong2024openess}

\textbf{Target motivation:} 

Event-based semantic segmentation (ESS) faces challenges due to the sparse, asynchronous, and high-temporal-resolution nature of event data, requiring expensive annotations for training. Existing methods are limited by closed-set learning and the difficulty of transferring knowledge from image to event domains. OpenESS addresses these limitations by proposing a novel framework that transfers pre-trained knowledge from image and text domains to event data, enabling open-vocabulary and zero-shot learning without dense event annotations.

\textbf{Summary of target experiment:} 

OpenESS introduces two key innovations: frame-to-event (F2E) contrastive distillation and text-to-event (T2E) consistency regularization, to facilitate effective cross-modality knowledge transfer from images and texts to event data for open-vocabulary semantic segmentation.

\textbf{Designs of target experiment:} 
\begin{itemize}
\item \textbf{Design 1:} 
   \begin{itemize}
   \item \textbf{Design name:}  Frame-to-Event (F2E) Contrastive Distillation
   \item \textbf{Description of design name:} This design leverages calibrated frames to generate superpixels and distills knowledge from a pre-trained image backbone to the event segmentation network. It uses a contrastive learning objective to transfer superpixel-level knowledge from dense frame pixels to sparse event streams, enhancing the event representation learning at higher granularity.
   \end{itemize}
\item \textbf{Design 2:} 
   \begin{itemize}
   \item \textbf{Design name:}  Text-to-Event (T2E) Consistency Regularization
   \item \textbf{Description of design name:} This design mitigates potential self-conflicts in the F2E contrastive learning by leveraging CLIP's text encoder to generate semantically consistent text-frame pairs. It constructs event-text pairs to encourage global-level alignment, using a semantic consistency regularization mechanism to improve the performance of open-vocabulary ESS.
   \end{itemize}
\item \textbf{Design 3:} 
   \begin{itemize}
   \item \textbf{Design name:} Thought-Augmented Reasoning
   \item \textbf{Description of design name:} For each problem, a relevant thought-template is retrieved from the meta-buffer and adaptively instantiated with specific reasoning structures to conduct efficient reasoning.
   \end{itemize}
\end{itemize}

\textbf{Topic:} 
The topic of this paper is open-vocabulary event-based semantic segmentation.

\textbf{Inspiration papers:} 
\begin{itemize}
\item Learning transferable visual models from natural language supervision~\cite{radford2021learning}
\item Ev-segnet: Semantic segmentation for event-based cameras~\cite{alonso2019ev}
\item Event-based vision: A survey~\cite{gallego2020event}
\item Ess: Learning event-based semantic segmentation from still images~\cite{sun2022ess}
\item SLIC superpixels compared to state-of-the-art superpixel methods~\cite{achanta2012slic}
\end{itemize}

\vspace{2\baselineskip}

\textbf{AI-Researcher:}
\begin{itemize}
\item \textbf{Motivation:} Existing approaches to open-vocabulary semantic segmentation often rely on augmenting event data with dense visual modalities like RGB images or pre-trained dense vision models. \sethlcolor{lightgreen}\hl{These methods, however, are resource-intensive and don't scale effectively for purely event-based data. Event cameras provide high temporal resolution but minimal spatial information, and their sparse, asynchronous nature makes direct pixel-level segmentation unreliable.} Humans, by contrast, use reasoning and context to 'fill in the blanks' in missing information. Inspired by this, \sethlcolor{lightgreen}\hl{we propose utilizing language models (LLMs) to iteratively reason about and predict missing spatial regions in sparse event data through a self-reconstruction framework.} This approach combines multimodal embeddings from temporally adjacent events with language-guided reasoning to improve region segmentation in an open-vocabulary setting.
\item \textbf{Experimental plan:} 
   \begin{itemize}
   \item \textbf{Proposed Method:} The proposed method introduces a self-reconstruction prompting pipeline via language models to predict and refine missing spatial regions in sparse event datasets. First, \sethlcolor{lightgreen}\hl{we segment ambiguous regions from event data.} These regions are then passed to a language model together with adjacent temporal events and multimodal embeddings, prompting the model to generate reconstruction hints such as \sethlcolor{lightred}\hl{'Predict the most probable shape and motion in these missing frames.' Intermediate reconstructions are iteratively refined by feedback prompts, such as 'Does this segment align with the trajectory of a running human?' Finally, the regions are mapped to open-vocabulary semantic labels using similarity-based alignment with pre-trained word embeddings (e.g., CLIP).}
   \item \textbf{Experiment step:}
        \begin{itemize}
               \item "Step 1: Gather Datasets": "We will use event-based datasets such as DDD17, N-Caltech101, and MVSEC. These datasets contain sparse event sequences with dynamic objects or scenes. Additionally, optical flow estimations from these datasets can provide temporal information for grounding reconstructions.",
                \item "Step 2: Preprocess Event Data": "Convert raw event streams into event frames by aggregating events over short time intervals. Additionally, compute optical flow or motion embeddings for adjacent events to encode temporal changes. Use an event-based simulator for testing performance under increasing sparsity levels.",
                \item "Step 3: Baseline Construction": "Implement two baselines: (1) direct event-based segmentation using a simple CNN; (2) event-to-RGB reconstruction models followed by segmentation with an open-vocabulary semantic segmentation system, such as MaskCLIP or CLIPSeg.",
                \item "Step 4: Construct Prompts for Reconstruction": "Design initial prompts to guide the LLM in predicting missing regions. Example prompts include:- 'Analyze the following sparse event patch. Predict the plausible geometry of the objects in the region and describe their semantic categories.'- 'Given the trajectory in these temporally adjacent embeddings, predict the continuation of object motion and shape.'- 'What semantic class is this shape most likely to belong to based on geometrical features?'",
                \item"Step 5: \sethlcolor{lightred}\hl{Iterative Self-Reconstruction Pipeline": "Use the following steps for iterative refinement:1. Initial reconstruction: Pass the event patch and temporally adjacent embeddings to the model. Let the LLM predict likely regions and classes.2. Feedback loop: Incorporate prompting to validate outputs. For example, 'Does this cyclic shape resemble a bicycle wheel? If not, refine the reconstruction.'3. Refine boundaries and motion predictions iteratively until convergence or consistent results.",}
                \item"Step 6: Map to Open-Vocabulary Classes": \sethlcolor{lightgreen}\hl{"Use pre-trained multimodal language-vision models like CLIP to align LLM predictions with open-vocabulary semantic categories. For example, compare representational embeddings of the predicted shape class ('bicycle') with candidate class embeddings from the event dataset."},
                \item"Step 7: Evaluate Metrics": "Evaluate segmentation performance using:- Mean Intersection over Union (MIoU): Measure segmentation accuracy across reconstructed regions and ground truth.- Semantic Adaptability: Quantify accuracy on unseen semantic categories (open-vocabulary performance).- Reconstruction Fidelity: Assess the quality of missing region predictions based on pixel or shape alignment with the ground truth.",
                \item "Step 8: Compare with Baselines": "Evaluate the performance of the proposed method against the baselines to quantify reconstruction improvements, open-vocabulary adaptability, and reduced ambiguities under sparse conditions."
        \end{itemize}
   \end{itemize}
\end{itemize}

\vspace{2\baselineskip}

\textbf{AI-Scientist:}
\begin{itemize}
\item \textbf{Motivation:} Event cameras offer unique advantages, such as high temporal resolution and dynamic range, but their sparsity and asynchronous nature present challenges for semantic segmentation. \sethlcolor{lightgreen}\hl{Existing methods either rely heavily on labeled datasets (Paper 2, Paper 4) or synthetic data (Paper 4), which limits scalability.} Open-vocabulary models like CLIP (Paper 1) have shown promise in performing zero-shot tasks by aligning image and text embeddings but have not been adapted to event-based segmentation. \sethlcolor{lightgreen}\hl{This research aims to address these gaps by proposing a cross-modal framework that aligns event embeddings with language embeddings}, enabling zero-shot and few-shot segmentation on event data.
\item \textbf{Experimental plan:} 
   \begin{itemize}
   \item \textbf{Experiment step:}
        \begin{itemize}
               \item  "Step1": "Adopt the 6-channel event representation (Paper 2) as the baseline. Encode spatial and temporal information into compact embeddings suitable for cross-modal tasks.",
                \item "Step2": "\sethlcolor{lightgreen}\hl{Train a cross-modal embedding model using contrastive loss to align event embeddings with text embeddings from a pre-trained model like CLIP.} Focus on zero-shot classification first as a preliminary evaluation task.",
                \item "Step3": "Extend the evaluation to zero-shot semantic segmentation by using text prompts for class names and comparing against existing baselines like ESS (Paper 4). Use datasets like DDD17 and DSEC-Semantic for testing.",
                \item"Step4": "Perform ablation studies to test the impact of the 6-channel representation versus alternative representations. Compare the performance of the proposed method against supervised event-based models (Paper 2) and UDA methods (Paper 4).",
                \item "Step5": "Explore few-shot fine-tuning using a small subset of labeled event data to optimize segmentation performance on specific domains or tasks, prioritizing practicality and generalization."
        \end{itemize}
   \end{itemize}
\end{itemize}

\vspace{2\baselineskip}
\textbf{VIRSC:}
\begin{itemize}
\item \textbf{Motivation:} This paper proposes a novel framework called Domain-Invariant Open-Vocabulary Event-Based Semantic Segmentation (DI-OVE-SS), which enhances the adaptability and robustness of event-based segmentation models by combining vision-language pretraining with domain-invariant event representations and hybrid learning techniques. The framework addresses two core challenges in the event-based vision space: (1) domain generalization under distribution shifts (e.g., varying lighting, environments, or noise levels) and (2) efficient adaptation to resource constraints in edge devices. DI-OVE-SS introduces a Transferable Event Representation (TER) module that learns domain-invariant features by disentangling temporal, spatial, and contextual information through self-supervised pretraining. Additionally, a hybrid training pipeline integrates supervised fine-tuning on small-scale labeled datasets with self-supervised pretraining on unlabeled event data, enabling effective learning in low-data regimes. The model leverages a lightweight modification of CLIP's architecture, adapting it for event streams through specialized embedding mechanisms. This framework not only enables zero-shot segmentation for unseen classes but also facilitates cross-domain transfer, making it suitable for real-world applications such as autonomous driving, robotics, and surveillance, where event streams vary significantly between environments. By advancing beyond existing approaches, DI-OVE-SS provides a scalable, efficient, and generalizable solution for open-vocabulary segmentation on event-based data.
\item \textbf{Experimental plan:} 
   \begin{itemize}
   \item \textbf{Experiment step:}
        \begin{itemize}
               \item "Step1": "Dataset Construction and Preprocessing: Extend existing event datasets (e.g., MVSEC, DDD17, and DVS128 Gesture) with additional domain-variant scenarios, such as day and night driving, indoor and outdoor environments, and high and low sensor noise. Use a combination of real-world event cameras and synthetic event data to simulate diverse conditions. Annotate a small portion of the dataset with semantic labels, while leaving the majority unlabeled for self-supervised pretraining.",
                \item  "Step2": "Transferable Event Representation (TER) Pretraining: Train a TER module using self-supervised learning objectives (e.g., contrastive loss with augmentation invariance) to disentangle spatial, temporal, and contextual features from event streams. The TER module should be designed to capture domain-invariant representations, ensuring robustness across different environmental conditions. Evaluate the quality of these representations using clustering and t-SNE visualization techniques.",
                \item  "Step3": "Hybrid Training: \sethlcolor{lightgreen}\hl{Fine-tune the vision-language framework (e.g., CLIP) on event streams using supervised training} on the small labeled subset of the dataset. Simultaneously, integrate the TER module into the architecture to provide domain-invariant inputs. Use a multitask learning objective combining supervised segmentation loss and self-supervised TER alignment loss to optimize the model.",
                \item "Step4": "Zero-Shot and Cross-Domain Evaluation: Test the model's zero-shot segmentation capabilities on unseen classes and domains without additional fine-tuning. Evaluate on tasks such as night driving or highly dynamic environments and benchmark against state-of-the-art event-based segmentation models. Use metrics such as mIoU, MIoU under domain shifts, and robustness to noise.",
                \item "Step5": "Resource Optimization and Edge Deployment: Evaluate the computational efficiency of the model by deploying it on edge devices (e.g., NVIDIA Jetson or Raspberry Pi). Test latency, energy consumption, and model size. Optimize the model further by pruning or quantization to fit the constraints of resource-constrained devices.",
                \item "Step6": "Ablation Study and Analysis: Perform detailed ablation studies to assess the contribution of the TER module, hybrid training objectives, and architectural modifications to CLIP. Analyze the impact of each component on segmentation accuracy, domain generalization, and computational efficiency. Validate the TER's robustness by visualizing gradients and feature maps across varying domains.",
                \item "Step7": "Scalability and Real-World Deployment: Test the model on larger datasets and complex real-world conditions to evaluate scalability. Experiment with new event camera sensors or synthetic data to simulate extreme scenarios like high-speed motion or low-light conditions. Validate the model's effectiveness in specific applications like autonomous driving or drone navigation." 
        \end{itemize}
   \end{itemize}
\end{itemize}

\vspace{2\baselineskip}
\textbf{SCIPIP}
\begin{itemize}
\item \textbf{Motivation:} Multimodal Event Alignment with Vision-Language Semantic Spaces  - **Details**:  This idea leverages \sethlcolor{lightred}\hl{**event-stream alignment with vision-language semantic spaces** to bridge the gap between sparse event representations and dense, pretrained image-text embeddings.} The core framework includes: 1. A **dedicated event encoder**: Converts event streams into feature embeddings that preserve temporal, geometric, and semantic properties.  2. A **multimodal projection layer**: Maps the encoded event representations into a shared semantic space of large vision-language models (e.g., CLIP, BLIP).   3. A **contrastive loss objective**: \sethlcolor{lightgreen}\hl{Ensures similarity between event embeddings, corresponding visual samples, and associated textual descriptions, enabling} **zero-shot generalization** to unseen vocabularies.  To strengthen this approach, a novel **temporal augmentation technique** is proposed, where noisy event streams are used to simulate realistic environments, improving the encoder robustness under real-world dynamic conditions. 
\end{itemize}

\end{tcolorbox}
\clearpage

\section{Case 1}
\label{appendix:Case_1}
\begin{tcolorbox}[breakable,enhanced,width=\textwidth]
\textbf{Target paper:} 

Buffer of Thoughts Thought-Augmented Reasoning with Large Language Models~\cite{yang2406buffer}

\textbf{Target motivation:} 

Current LLMs have shown impressive performance in reasoning tasks, but existing prompting methods face limitations in universality, computational intensity, and lack of generalizable guidelines from past tasks. BoT addresses these by providing a scalable and stable framework that leverages historical insights for improved reasoning.

\textbf{Summary of target experiment:} 

BoT introduces a meta-buffer to store high-level thoughts (thought-templates) and a buffer-manager to dynamically update the meta-buffer, enabling adaptive instantiation of reasoning structures for efficient problem-solving.

\textbf{Designs of target experiment:} 
\begin{itemize}
\item \textbf{Design 1:} 
   \begin{itemize}
   \item \textbf{Design name:}  Meta-Buffer
   \item \textbf{Description of design name:} A lightweight library that stores high-level thoughts (thought-templates) distilled from various problem-solving processes, allowing for shared reasoning structures across tasks.
   \end{itemize}
\item \textbf{Design 2:} 
   \begin{itemize}
   \item \textbf{Design name:}  Buffer-Manager
   \item \textbf{Description of design name:} A component that dynamically updates the meta-buffer by summarizing high-level guidelines and thoughts from each problem-solving process, enhancing the meta-buffer's capacity as more tasks are solved.
   \end{itemize}
\item \textbf{Design 3:} 
   \begin{itemize}
   \item \textbf{Design name:} Thought-Augmented Reasoning
   \item \textbf{Description of design name:} For each problem, a relevant thought-template is retrieved from the meta-buffer and adaptively instantiated with specific reasoning structures to conduct efficient reasoning.
   \end{itemize}
\end{itemize}

\textbf{Topic:} 
The topic of this paper is enhancing reasoning in large language models.

\textbf{Inspiration papers:} 
\begin{itemize}
\item Chain-of-thought prompting elicits reasoning in large language models~\cite{wei2022chain}
\item Tree of thoughts: Deliberate problem solving with large language models~\cite{yao2023tree}
\item Expertprompting: Instructing large language models to be distinguished experts~\cite{xu2023expertprompting}
\item Pal: Program-aided language models~\cite{gao2023pal}
\item Self consistency improves chain of thought reasoning in language models~\cite{wang2022self}
\end{itemize}

\vspace{2\baselineskip}

\textbf{AI-Researcher:}
\begin{itemize}
\item \textbf{Motivation:} Existing reasoning techniques, such as Chain-of-Thought prompting or ensemble methods, rely on sequential prompts that guide models along a single rationale pathway. These methods fail to incorporate insights \sethlcolor{lightgreen}\hl{across multiple reasoning streams, resulting in suboptimal solutions for problems that require balancing diverse perspectives}. Humans excel at reasoning by considering multiple rationale streams simultaneously and synthesizing cross-perspective insights to form synergistic conclusions. Inspired by this human capability, \sethlcolor{lightgreen}\hl{we propose Multi-Rationale Synergy Prompting (MRSP), which prompts LLMs to generate independent reasoning pathways, critique insights across streams, and synthesize solutions accounting for trade-offs}.
\item \textbf{Experimental plan:} 
   \begin{itemize}
   \item \textbf{Proposed Method:} Multi-Rationale Synergy Prompting involves three distinct steps: 1. **Generate Independent Rationales**: \sethlcolor{lightgreen}\hl{The LLM is prompted to generate reasoning chains independently for each aspect or perspective of the task (e.g., ethical reasoning and efficiency reasoning).} 2. **Critique and Cross-Reference**: \sethlcolor{lightgreen}\hl{The LLM critiques the insights from each perspective by identifying points of agreement, conflict, or tension between them.} 3. **Synergized Conclusion**: The LLM synthesizes findings from the critiques, creating a unified solution that balances the trade-offs and incorporates strengths from all rationale streams. Prompts include commands such as, "Generate independent rationales for Perspective A and Perspective B," followed by "Critique and identify the points of synergy and tension between these rationales," and finally "Generate a unified answer considering all rationales.
   \item \textbf{Experiment step:}
        \begin{itemize}
               \item "Step 1: Gather Datasets": "Select benchmarking datasets that require reasoning across multiple perspectives:- Moral Machines: Contains ethical dilemmas for autonomous vehicles requiring consideration of human values.- Ethical Dilemmas QA: Requires reasoning through competing ethical priorities.- Social IQa: Involves questions related to social commonsense, requiring nuanced trade-offs.- Create synthetic examples if necessary to evaluate tasks requiring synergy across diverse perspectives (e.g., balancing financial, ethical, and operational priorities)."
                \item "Step 2: Construct Prompts": "Design prompts following the MRSP methodology:- **Independent Rationales Prompt**: Example: 'Generate reasoning chains for both ethical justification and operational efficiency of [problem statement].'- **Critique Prompt**: Example: 'Critique the reasoning chains you generated earlier for ethical and operational perspectives. Highlight points of commonality, tension, or disagreement.'- \sethlcolor{lightred}\hl{**Synthesis Prompt**: Example: 'Based on the critique, synthesize a solution that balances ethical concerns with operational priorities while ensuring alignment with the original task.'",}
                \item"Step 3: Baseline Design": "Compare MRSP against three baselines:- **Sequential Chain-of-Thought Prompting (CoT)**: Generate reasoning step-by-step along a single pathway.- **Direct Prompting**: Provide the task directly to the model without intermediate rationales.- **Ensemble Prompting**: Generate reasoning streams separately but average their outputs without critique or synthesis.",
                \item"Step 4: Select Models": "Test MRSP and baselines using multiple LLMs:- **Black-box APIs**: {GPT-3.5 (Text-Davinci-003), GPT-4, Claude 2, and Gemini.- **Open-Source Models**: LLaMA-2-70B-chat} and similar models for comparison.Ensure prompts are compatible with model capabilities and consistent across platforms.",
                \item"Step 5: Execution": "Run the experiments:1. Input each dataset example into the LLM using baseline methods (CoT, Direct prompting, Ensemble prompting).2. Input each dataset example using MRSP (multi-step prompts for independent rationales, critique, and synthesis).3. Collect outputs from all methods for analysis.",
                \item"Step 6: Evaluation Metrics": "Measure LLM performance along three criteria:- **Coherence**: Are synthesized solutions logically connected and comprehensible?- **Thoroughness**: Do solutions incorporate reasoning from all provided perspectives?- **Trade-Off Balancing**: How well does the synthesis balance competing priorities from different rationale streams?Use automatic metrics (e.g., BLEU/ROUGE for text overlap) and task-specific scoring (e.g., correctness and depth for reasoning tasks).",
                \item"Step 7: Analyze Results": "Compare MRSP outputs to baselines:1. Compare overall task accuracy and coherence across methods.2. Analyze model consistency in synthesizing solutions.3. Generate detailed performance breakdowns for each dataset and reasoning dimension (ethical, operational, commonsense reasoning)."
        \end{itemize}
   \end{itemize}
\end{itemize}

\vspace{2\baselineskip}

\textbf{AI-Scientist:}
\begin{itemize}
\item \textbf{Motivation:} While existing methods like CoT and ToT significantly improve reasoning by structuring intermediate steps, \sethlcolor{lightgreen}\hl{they lack mechanisms to simulate diverse perspectives or incorporate iterative critique into the reasoning process.} Real-world human reasoning often involves collaborative problem-solving where multiple agents (or perspectives) debate, refine, and reconcile their thoughts. This paper aims to fill this gap by introducing a framework that \sethlcolor{lightgreen}\hl{enables LLMs to simulate multiple reasoning agents, fostering diversity and robustness in reasoning paths.}
\item \textbf{Experimental plan:} 
   \begin{itemize}
   \item \textbf{Experiment step:}
        \begin{itemize}
               \item "Step1": \sethlcolor{lightgreen}\hl{"Enable the LLM to simulate multiple reasoning agents by generating diverse reasoning paths for the same problem. Each path reflects an independent reasoning perspective.",}
                \item "Step2": \sethlcolor{lightred}\hl{"Introduce a critique phase where the reasoning paths are cross-examined. Each reasoning path critiques others by highlighting inconsistencies, errors, or gaps."},
                \item "Step3": "\sethlcolor{lightgreen}\hl{Implement a simple reconciliation mechanism where the final answer is chosen based on majority voting among the reasoning paths that remain after critique. This ensures robustness without added complexity.}",
                \item"Step4": "Evaluate the CRF on reasoning benchmarks (e.g., GSM8K, AQuA, StrategyQA) and compare its performance to CoT, ToT, PAL, and self-consistency. Analyze both accuracy improvements and the diversity of reasoning paths.",
                \item "Step5": "Perform qualitative analysis by examining examples where CRF succeeds or fails, emphasizing the interpretability and robustness of its multi-agent reasoning process."
        \end{itemize}
   \end{itemize}
\end{itemize}

\vspace{2\baselineskip}
\textbf{VIRSC:}
\begin{itemize}
\item \textbf{Motivation:} The proposed framework, 'Hybrid Adaptive Reflective System with External Symbolic Computation (HARS-ESC),' combines adaptive reasoning with external symbolic computation tools to enhance robustness and adaptability in reasoning tasks for large language models (LLMs). HARS-ESC integrates a meta-cognitive layer inspired by ARToT, which evaluates task complexity, uncertainty, and ambiguity. \sethlcolor{lightred}\hl{This layer dynamically determines whether the task can be solved internally through adaptive reflective reasoning or whether it requires external computation (e.g., symbolic interpreters, solvers, or APIs).} For example, computationally intensive or error-prone steps such as arithmetic or symbolic manipulation can be offloaded to external systems, while nuanced or abstract reasoning is handled internally. The system utilizes a modular architecture to seamlessly switch between internal and external reasoning modes based on task demands. \sethlcolor{lightred}\hl{Additionally, HARS-ESC introduces a 'task-decomposition planner' that divides problems into discrete sub-tasks, assigns each sub-task to the most suitable reasoning module (internal or external), and evaluates intermediate outputs for logical consistency and task alignment.} The framework also incorporates a feedback-driven mechanism to iteratively refine both the task decomposition and the reasoning approach. This ensures continuous improvement and minimizes logical errors or inefficiencies. HARS-ESC is designed for diverse applications like multi-step scientific reasoning, adaptive learning systems, real-time decision-making, and tasks involving incomplete or ambiguous data. By leveraging both adaptive strategies and external computation, HARS-ESC offers improved accuracy, resource efficiency, and robustness across a wide range of reasoning tasks.
\item \textbf{Experimental plan:} 
   \begin{itemize}
   \item \textbf{Experiment step:}
        \begin{itemize}
               \item "Step1": "\sethlcolor{lightred}\hl{Develop the meta-cognitive layer to evaluate task complexity and ambiguity. Enhance the layer to include a 'diversity trigger,' which determines whether diverse reasoning paths should be generated for the given task based on a set of heuristics (e.g., high ambiguity, incomplete data)."},
                \item  "Step2": "Design and implement the diversity-aware reasoning module. Use probabilistic sampling techniques (e.g., temperature sampling, nucleus sampling) and modular task decomposition to generate diverse reasoning paths for each sub-task. Leverage both neural internal reasoning and external symbolic computation based on the nature of each sub-task.",
                \item  "Step3": "Develop the reconciliation engine. For fixed-answer tasks, implement marginalization techniques to consolidate the outputs of diverse reasoning paths. For open-ended tasks, use semantic alignment and clustering methods to synthesize a unified response from divergent outputs.",
                \item "Step4": "Integrate the meta-cognitive layer, diversity-aware reasoning module, and reconciliation engine into a cohesive pipeline. Ensure seamless switching between internal and external reasoning and effective communication between modules.",
                \item "Step5": "Evaluate DAHRS on a range of benchmarks, including (a) fixed-answer multi-step reasoning datasets like GSM8K, StrategyQA, and ARC-Challenge, (b) open-ended tasks requiring creative reasoning (e.g., scientific hypothesis generation, ethical dilemmas), and (c) real-time planning scenarios with incomplete or ambiguous data inputs.",
                \item "Step6": "Compare DAHRS against HARS-ESC, self-consistency, and existing baseline methods like chain-of-thought (CoT) prompting, PAL, and RToT. Use metrics such as reasoning accuracy, robustness to ambiguity, computational efficiency, and reasoning diversity to measure performance.",
                \item "Step7": "Conduct qualitative analyses to study how reasoning diversity impacts task outcomes. Analyze whether the reconciliation engine effectively synthesizes diverse reasoning paths into a coherent and interpretable output.",
                \item "Step8": "Test DAHRS in resource-constrained environments by scaling down the model size and observing the trade-offs between accuracy, computational cost, and interpretability. Validate that the diversity-aware mechanism scales well with reduced resources."         
        \end{itemize}
   \end{itemize}
\end{itemize}

\vspace{2\baselineskip}
\textbf{SCIPIP}
\begin{itemize}
\item \textbf{Motivation:}  **Hierarchical Neuro-Symbolic Reasoning with Modular Task Decomposition (HNSR)**  - **Details**: Elevate the neuro-symbolic hybrid framework by incorporating modular task decomposition with a scalable hierarchical architecture. \sethlcolor{lightred}\hl{In this framework, reasoning problems are systematically decomposed into **symbolically hard** (e.g., formal proofs, combinatorial optimization) and **neuro-adaptive** tasks (e.g., intuitive reasoning, complex pattern recognition).} A controller module leverages symbolic reasoning for problem decomposition and assigns subtasks to the symbolic or neural processing units based on complexity and cognitive type. Moreover, the system generates **transparent reasoning blueprints** that document subtask allocations, decision pathways, and results. \sethlcolor{lightred}\hl{Dynamic complexity redistribution is guided by real-time analysis of performance metrics, such as reasoning latency and subtask error correction rates. Feedback loops further refine modular performance for future challenges.}

\end{itemize}

\end{tcolorbox}
\clearpage

\section{Case 2}
\label{appendix:Case_2}
\begin{tcolorbox}[breakable,enhanced,width=\textwidth]
\textbf{Target paper:} 

MDAgents: An Adaptive Collaboration of LLMs for Medical Decision-Making~\cite{yang2406buffer}

\textbf{Target motivation:} 

Despite the potential of LLMs in medical applications, there is a lack of effective frameworks that integrate the real-world systematic Medical Decision-Making (MDM) process, which requires an adaptive, collaborative, and tiered approach. MDAgents addresses this gap by emulating the hierarchical diagnosis procedures ranging from individual clinicians to collaborative clinician teams, aiming to improve accuracy and efficiency in medical tasks.

\textbf{Summary of target experiment:} 

MDAgents framework incorporates four stages: Medical Complexity Check, Expert Recruitment, Analysis and Synthesis, and Decision-making. It dynamically assigns LLMs to roles independently or within groups based on the task's complexity, utilizing prompting techniques and collaborative strategies to enhance decision-making.

\textbf{Designs of target experiment:} 
\begin{itemize}
\item \textbf{Design 1:} 
   \begin{itemize}
   \item \textbf{Design name:}  Medical Complexity Check
   \item \textbf{Description of design name:} The system evaluates the medical query, categorizing it as low, moderate, or high complexity based on clinical decision-making techniques. This step ensures that the complexity level of the query is accurately assessed to determine the appropriate collaboration structure.
   \end{itemize}
\item \textbf{Design 2:} 
   \begin{itemize}
   \item \textbf{Design name:}  Expert Recruitment
   \item \textbf{Description of design name:} Based on the complexity assessment, the framework activates a single Primary Care Clinician (PCC) for low complexity issues, or a Multi-disciplinary Team (MDT) or Integrated Care Team (ICT) for moderate or high complexities. This ensures that the right expertise is engaged for each query.
   \end{itemize}
\item \textbf{Design 3:} 
   \begin{itemize}
   \item \textbf{Design name:} Analysis and Synthesis
   \item \textbf{Description of design name:} For low complexity queries, solo agents use prompting techniques like Chain-of-Thought (CoT) and Self-Consistency (SC). For moderate and high complexities, MDTs and ICTs involve multiple LLM agents forming a consensus or synthesizing information, respectively, to address the query comprehensively.
   \end{itemize}
\item \textbf{Design 4:} 
   \begin{itemize}
   \item \textbf{Design name:} Decision-making
   \item \textbf{Description of design name:} The final stage synthesizes all inputs to provide a well-informed answer to the medical query. It employs ensemble techniques to ensure the decision is robust and reflects a consensus among the models when applicable.
   \end{itemize}
\end{itemize}

\textbf{Topic:} 
The topic of this paper is enhancing the utility of Large Language Models in medical decision-making.

\textbf{Inspiration papers:} 
\begin{itemize}
\item Large language models are few-shot clinical information extractors~\cite{agrawal2022large}
\item NLICE: Synthetic Medical Record Generation for Effective Primary Healthcare Differential Diagnosis~\cite{al2023nlice}
\item Ehrxqa: A multi-modal question answering dataset for electronic health records with chest x-ray images~\cite{bae2023ehrxqa}
\item Comparative accuracy of diagnosis by collective intelligence of multiple physicians vs individual physicians~\cite{barnett2019comparative}
\item The use of analytic hierarchy process for measuring the complexity of medical diagnosis~\cite{ben2020use}
\end{itemize}

\vspace{2\baselineskip}

\textbf{AI-Researcher:}
\begin{itemize}
\item \textbf{Motivation:} Existing temporal reasoning methods in medical AI rely heavily on structured representations like timelines or domain-specific models. \sethlcolor{lightred}\hl{While effective in some cases, these approaches are rigid, require extensive domain knowledge, and do not leverage recent advances in general-purpose large language models.} LLMs offer the potential to process free-text medical records, making them accessible and flexible. However, their ability to reason temporally in free-text settings is currently limited. Our inspiration comes from the idea that guiding LLMs using temporal reasoning prompts can simulate the ability to interpret and analyze patient histories in a temporally grounded manner. By breaking problems into temporally ordered steps, we hypothesize that LLMs can significantly improve their diagnostic accuracy compared to conventional prompting methods.
\item \textbf{Experimental plan:} 
   \begin{itemize}
   \item \textbf{Proposed Method:} We propose a novel framework called 'Temporal Reasoning Prompting' (TRP), which introduces a structured three-step prompting protocol for LLMs: 1. **Temporal Decomposition:** Analyze and organize the presented patient history into a sequence of temporally ordered events using a prompt like, 'Break down this patient's history into temporally ordered events, including symptom onset, progression, and any interventions. \sethlcolor{lightred}\hl{'2. **Temporal Reasoning:** Reason about the temporal sequence of events to identify plausible diagnoses using a prompt like, 'Considering the order and timing of the events listed, what diagnoses could fit this pattern?'3. **Validation and Iteration:** Validate assumptions and explore alternatives using iterative prompts such as, 'Could [event A] have occurred as a result of [event B]?' or 'Are there alternative sequences that could explain [event outcome]?}
   \item \textbf{Experiment step:}
        \begin{itemize}
               \item "Step 1: Gather Datasets": "We use synthetic patient data from the Synthea dataset, which generates realistic, time-stamped medical records, to \sethlcolor{lightred}\hl{create benchmark tasks for temporal reasoning.} Additionally, we construct synthetic test scenarios with predefined temporal patterns for diseases such as measles, Lyme disease, and COVID-19. For example, 'Patient develops fever 48 hours before a rash.'",
                \item "Step 2: Define Baselines and Evaluation Metrics"
                \begin{itemize}
                    \item  "Baselines": "We consider two baselines: (1) direct prompting (flat question-answering, no temporal decomposition); (2) few-shot chain-of-thought (CoT) prompting without temporal reasoning focus. For direct prompting, we ask the model for a diagnosis directly. For CoT prompting, we use 'Let's think step by step' before the diagnostic question.",
                    \item "Metrics":
                    \begin{enumerate}
                        \item "Diagnostic Accuracy": "Percentage of correct diagnoses.",
                        \item "Temporal Consistency Score": "Measures how often outputs conform to temporal constraints in the input data.",
                        \item "Explainability Assessment": "Qualitative evaluation of the intermediate steps."
                    \end{enumerate}   
                \end{itemize}
                \item"Step 3: Construct Temporal Reasoning Prompts": 
                \begin{itemize}
                    \item "Examples of Prompts for Temporal Decomposition":
                                "Input: 'The patient developed a headache on Tuesday, vomiting on Thursday, and sleep disturbances starting last week after a trip to a rural area.' Prompt: 'Break down this patient's history into a temporally ordered list of events with dates or time markers.'Expected Output: '1. Sleep disturbances (last week); 2. Headache (Tuesday); 3. Vomiting (Thursday).'"
                        \item \sethlcolor{lightred}\hl{"Examples of Prompts for Temporal Reasoning": 
                                "Input: 'Consider the following events: 1. Fever on January 1st; 2. Rash on January 3rd. Prompt: 'Considering the timing and order of events, suggest possible diagnoses supported by this sequence.'Expected Output: 'Possible diagnoses: 1. Measles (fever typically precedes rash by 2-4 days); 2. Chickenpox (fever can precede rash by 1-3 days).' "}
                            \item "Examples of Prompts for Validation":
                                "Input: 'Event sequence: 1. Patient develops high fever; 2. Patient experiences hallucinations two days later. Diagnosis: Brain infection.' Prompt: 'Could the hallucinations have been a result of the high fever? Are there alternative diagnoses possible for this sequence?'Expected Output: 'Yes, hallucinations could result from high fever (febrile delirium). Alternative diagnoses might include meningitis, encephalitis.'"
                \end{itemize}
                \item"Step 4: Model Selection and Configuration": 
                \begin{itemize}
                    \item "Models": {"We use GPT-3.5 and GPT-4 from OpenAI via API access for black-box evaluation. Optionally, we include LLaMA-2-70B-chat for comparative results."},
                    \item  "Configurations": "Set temperature to 0.7 for more creative outputs during reasoning and decomposition, and 0.5 for validation to focus on accuracy."
                \end{itemize}
                \item"Step 5: Run Experiments": 
                \begin{itemize}
                    \item "Experiment 1: Diagnostic Accuracy": "Compare the diagnostic accuracy of the baseline methods (direct prompting, chain-of-thought prompting) against TRP on Synthea synthetic patient records. Measure how many diagnoses match the correct answers.",
                    \item  "Experiment 2: Temporal Consistency": "Test whether the generated outputs adhere to temporal constraints in predefined synthetic cases (e.g., 'fever must precede rash by 48 hours'). Report the temporal consistency score.",
                    \item  "Experiment 3: Ablation Study": "Test the effect of removing or altering one step in the TRP framework (e.g., skipping the validation stage) and evaluate its impact on performance metrics.",
                    \item  "Experiment 4: Explainability Assessment": "Qualitatively compare the intermediate reasoning steps generated by the TRP prompts versus CoT and direct prompting baselines."
                \end{itemize}
                \item "Step 6: Analyze Results": "Visualize and compare the diagnostic accuracy and temporal consistency scores of different methods in tabular and graphical formats. Perform a qualitative analysis of the explainability and interpretability of the model's outputs."
        \end{itemize}
   \end{itemize}
\end{itemize}

\vspace{2\baselineskip}

\textbf{AI-Scientist:}
\begin{itemize}
\item \textbf{Motivation:} While LLMs have shown promise in tasks like clinical text analysis and question answering, and collective intelligence among humans has proven to improve diagnostic accuracy, the combination of LLM outputs with human expertise remains unexplored. \sethlcolor{lightgreen}\hl{Existing research does not address the optimal way to combine these two sources of intelligence in a scalable, practical manner. This study aims to bridge that gap by proposing a collaboration framework between LLMs and human experts for medical decision-making.}
\item \textbf{Experimental plan:} 
   \begin{itemize}
   \item \textbf{Experiment step:}
        \begin{itemize}
               \item "Step1": "Task Design - Select tasks like differential diagnosis, clinical correlation identification, and medication recommendations. Use publicly available datasets such as MIMIC-IV or synthetic datasets aligned with previous studies.",
                \item  "Step2": "LLM Output Generation - {Use a pre-existing LLM (e.g., GPT-3) to generate ranked outputs for the selected tasks.} Outputs will include ranked lists of possible diagnoses, symptoms, or decisions.",
                \item  "Step3": \sethlcolor{lightgreen}\hl{"Human Assessment - Engage participants with varying levels of medical expertise (e.g., medical students, residents, and experienced physicians) to individually rank their own outputs for the same tasks.} Use datasets with known answers to evaluate their baseline accuracy.",
                \item  "Step4": "Collaboration Mechanism - Implement simplified aggregation strategies, such as weighted averaging, to combine LLM outputs with human inputs. Vary the weight given to LLMs versus humans (e.g., 80-20, 50-50 ratios) and compare how these strategies perform across levels of expertise.",
                \item "Step5": "Evaluation - Compare the diagnostic accuracy and consistency of LLM-human collaborative outputs with standalone LLM and human outputs. Use metrics like Top-1 and Top-5 accuracy, precision, and recall.",
                \item "Step6": "Analysis - Investigate how human expertise levels and weighting strategies influence the performance of the collaboration. Identify conditions where collaboration provides the most significant improvements.
        \end{itemize}
   \end{itemize}
\end{itemize}

\vspace{2\baselineskip}
\textbf{VIRSC:}
\begin{itemize}
\item \textbf{Motivation:} The proposed framework, titled 'MedCollective-GPT: A Collective Intelligence-Enhanced Framework for Medical Decision-Making,' introduces an innovative paradigm where multiple instances of large language models (LLMs) collaborate with clinicians to improve diagnostic accuracy and decision-making in healthcare settings. This approach integrates collective intelligence principles with LLM capabilities to create a more robust, scalable, and interpretable system for handling complex medical cases. The system comprises three core components: \sethlcolor{lightgreen}\hl{(1) Multi-Agent LLM Collaboration: Multiple specialized LLM instances (fine-tuned on different medical domains, e.g., radiology, pathology, and general medicine) generate independent diagnostic recommendations that are aggregated using a novel proportional weighting rule inspired by collective intelligence. Each LLM agent provides insights reflecting its domain expertise, reducing the likelihood of domain-specific biases; (2) Human-in-the-Loop Coordination: Clinicians interact with the aggregated recommendations to validate or refine diagnostic hypotheses. The system incorporates clinician feedback as reinforcement signals to dynamically improve the weighting mechanism and ensure relevance to real-world settings;} (3) Adaptive Explainability and Causal Reasoning: The framework provides interpretable insights by generating explanations for why certain diagnoses or treatment options were prioritized. It also leverages causal inference to simulate 'what-if' scenarios, predicting potential outcomes of medical interventions. This integration ensures that the system not only provides accurate predictions but also supports clinicians in understanding the reasoning process. The novelty lies in combining LLM-driven reasoning with collective intelligence mechanisms, enabling synergistic decision-making in complex, high-stakes scenarios. Ethical principles such as differential privacy, audit trails, and fairness are embedded into the system to ensure its safe and ethical deployment in clinical environments.
\item \textbf{Experimental plan:} 
   \begin{itemize}
   \item \textbf{Experiment step:}
        \begin{itemize}
               \item "Step 1": "Dataset Construction: Create a cohesive multi-modal medical dataset integrating structured EHR data, imaging data (e.g., chest X-rays, MRIs), and time-series patient-specific data. This dataset will be augmented by domain-specific annotations for differential diagnoses and causal reasoning to stress-test the collective intelligence mechanism.",
                \item   "Step 2": "Multi-Agent LLM Training: \sethlcolor{lightgreen}\hl{Train multiple specialized LLMs using diverse medical datasets. Fine-tune each LLM on specific domains (e.g., oncology, cardiology) for domain-specific expertise. Each LLM will independently generate diagnostic recommendations for a shared set of cases."},
                \item   "Step 3": "Proportional Weighting Aggregation: Develop and integrate a proportional weighting mechanism inspired by Paper 4. \sethlcolor{lightgreen}\hl{This mechanism will aggregate the outputs of the multi-agent LLMs by assigning higher weights to primary recommendations and reducing the influence of conflicting secondary diagnoses. The mechanism will dynamically adapt based on the performance of each agent over time."},
                \item  "Step 4": "Human-in-the-Loop Feedback: Implement a feedback loop where clinicians validate or refine the aggregated recommendations. Feedback will be used to fine-tune the proportional weighting rule and improve the system's domain-specific reasoning capabilities through reinforcement learning.",
                \item  "Step 5": "Explainability and Causal Reasoning Module: Develop an explainability module where each LLM instance and the aggregated system provide rationale for their diagnoses. Additionally, integrate a causal reasoning system to simulate 'what-if' scenarios, allowing clinicians to explore potential outcomes of various treatment options.",
                \item "Step 6": "Ethical Safeguards: Embed differential privacy mechanisms for data security and audit trail systems to maintain transparency in the decision-making process. Ensure fairness by testing the system's performance on diverse patient demographics and pathology types.",
                \item  "Step 7": "Evaluation and Benchmarking: Evaluate the system on a synthetic and real-world dataset, benchmarking its performance against state-of-the-art models like GPT-4 and MedFusion-RLX. Metrics will include Top-1 diagnostic accuracy, diagnostic diversity, explainability quality, and clinician trust measured through usability surveys.",
                \item  "Step 8": "Clinical Validation: Deploy the system in a controlled clinical environment, focusing on complex cases requiring input from multiple specialties (e.g., oncology, radiology, and cardiology). Monitor the system's performance, adaptability, and clinician acceptance over an extended period."         
        \end{itemize}
   \end{itemize}
\end{itemize}

\vspace{2\baselineskip}
\textbf{SCIPIP}
\begin{itemize}
\item \textbf{Motivation:} Symptom-Condition Complexity Fusion Model**- **Details**: \sethlcolor{lightred}\hl{Introduce a **dual-stream architecture** for processing patient data with one stream focused on unstructured natural language (patient symptoms, clinical notes) and the other on structured complexity signals (e.g., rarity, outcome severity, comorbidity risks).} The architecture includes: 1. **Fusion Layer**: Combines unsupervised embeddings from the natural language stream with complexity scores from the structured data stream. This provides a **weighted joint-representation** of patient data, where complexity-sensitive signals take priority in decision-making under ambiguous conditions. 2. **Recalibration Module**: Reranks outputs based on complexity thresholds. For instance, if multiple diagnostic suggestions share similar probabilities, prioritize those involving higher-complexity conditions. - **Enhancement**: Use a hybrid deep learning model, where the unstructured stream employs transformers and the structured stream leverages graph neural networks (GNNs) to model intricate patient-state relationships. Regularize the fusion layer to avoid overfitting to either simple or complex cases.  - **Output**: Enhances the ability of LLMs to interpret ambiguous patient symptoms in the context of high-complexity medical conditions while retaining fluency in natural language interactions.
\end{itemize}
\end{tcolorbox}
\clearpage

\section{Case 3}
\label{appendix:Case_3}
\begin{tcolorbox}[breakable,enhanced,width=\textwidth]
\textbf{Target paper:} 

Koala: Key frame-conditioned long video-llm~\cite{tan2024koala}

\textbf{Target motivation:} 

Existing vLLMs struggle with understanding long videos due to the loss of fine-grained spatiotemporal information when extracting key frames at a coarse rate. This limitation hinders their performance on complex and long-term temporal understanding tasks. Koala addresses this by introducing a novel approach that conditions the LLM on key frames to aggregate spatiotemporal context over longer temporal horizons, aiming to improve long video understanding.

\textbf{Summary of target experiment:} 

Koala introduces Conditioned Segment (CS) and Conditioned Video (CV) tokenizer functions that leverage learnable spatiotemporal queries to adapt the frozen video tokenizer function in pretrained vLLMs for long video understanding.

\textbf{Designs of target experiment:} 
\begin{itemize}
\item \textbf{Design 1:} 
   \begin{itemize}
   \item \textbf{Design name:}  Conditioned Segment Tokenizer
   \item \textbf{Description of design name:} This tokenizer repurposes the key frames tokenizer to select important frame-level information pertinent to both the local context of each segment and the global context of the key frames tokens. It modifies the video QFormer to condition on key frame tokens when aggregating input video segment features, ensuring compatibility with the video QFormer through learnable queries.
   \end{itemize}
\item \textbf{Design 2:} 
   \begin{itemize}
   \item \textbf{Design name:}  Conditioned Video Tokenizer
   \item \textbf{Description of design name:} This tokenizer reasons about important spatiotemporal relationships across segments, modeling how local segments are related to each other conditioned on the global context of the entire video. It introduces learnable temporal and concept queries to adapt segment tokens for compatibility with the video QFormer, facilitating the dissemination of global video context to more fine-grained visual concepts.
   \end{itemize}
\end{itemize}

\textbf{Topic:} 
The topic of this paper is improving long video understanding in video-Large Language Models (vLLMs).

\textbf{Inspiration papers:} 
\begin{itemize}
\item Videochat: Chat-centric video understanding~\cite{li2023videochat}
\item Egoschema: A diagnostic benchmark for very long-form video language understanding~\cite{mangalam2023egoschema}
\item Howto100m: Learning a text-video embedding by watching hundred million narrated video clips~\cite{miech2019howto100m}
\item Video-llama: An instruction-tuned audio-visual language model for video understanding~\cite{zhang2023video}
\item Seed-bench: Benchmarking multimodal llms with generative comprehension~\cite{wang2023fine}
\end{itemize}

\vspace{2\baselineskip}

\textbf{AI-Researcher:}
\begin{itemize}
\item \textbf{Motivation:} Existing methods like VideoChat-Embed and mPLUG-Owl attempt to process long videos but fail for two key reasons: (1) dense embeddings often lead to loss of crucial long-term contextual dependencies, (2) reasoning over such long contexts is computationally expensive and challenging for current vLLMs. Inspired by how humans process long narratives—by segmenting them into smaller, coherent chunks for easier understanding—we propose a novel approach called Temporal-Decomposition Prompting (TDP). \sethlcolor{lightgreen}\hl{By breaking videos into smaller, manageable temporal segments and reasoning step-by-step}, we aim to preserve coherence and enable improved long-term reasoning. \sethlcolor{lightgreen}\hl{TDP explicitly guides vLLMs to focus on key parts of the video iteratively}, enabling better temporal inference and spatiotemporal alignment. This method avoids computational overload while resolving deficiencies in existing baselines.
\item \textbf{Experimental plan:} 
   \begin{itemize}
   \item \textbf{Proposed Method:} \sethlcolor{lightred}\hl{Temporal-Decomposition Prompting (TDP) is designed to segment, summarize, and reason over long videos iteratively.} Here's a breakdown of how the method works:\sethlcolor{lightgreen}\hl{1. **Temporal Segmentation**: Using video metadata or scene detection tools (e.g., shot boundary detection algorithms, heuristic segmentation every 30 seconds), the video is divided into temporally coherent segments.2. **Segment Summarization**: For each segment, the vLLM is prompted to summarize the segment's key spatiotemporal details (e.g., Summarize Segment X What events occur in this timeframe?).3. **Reasoning Across Segments**: The summaries from all segments are stitched together and aligned for higher-level reasoning. Prompts guide the model to perform complex tasks such as event alignment, cause-effect reasoning, and temporal ordering using phrases such as Combine events from Segment X and Segment Y and infer their relationship.} 4. **Iterative Refinement**: If needed, the model revisits segments and updates reasoning outputs for improved coherence and consistency. For example, the model is guided with iterative prompts like, Analyze any inconsistencies or unclear relations between summaries and resolve them.With this methodology, \sethlcolor{lightred}\hl{we ensure the model processes videos incrementally, focusing on one part of the video at a time while continuously maintaining long-term coherence.}
   \item \textbf{Experiment step:}
        \begin{itemize}
               \item "Step 1: Gather Datasets": "We will use datasets that span complex long-term video contexts requiring fine-grained temporal understanding, such as Ego4D (for egocentric long video tasks) and VidSitu (for spatiotemporal understanding). Select subsets of the data that involve video clips lasting several minutes and require temporal reasoning capabilities.",
                \item  \sethlcolor{lightgreen}\hl{"Step 2: Implement Temporal Segmentation": "Use automated scene detection algorithms like PySceneDetect or temporal heuristics (e.g., divide videos into 30-second clips) to represent coherent segments. Annotate each segment with time boundaries.Example implementation:- Input video: 'walking\_dog.mp4'- Output: Segment timestamps [(0s-30s), (30s-60s), (60s-90s)...]"},
                \item"Step 3: Construct Baseline Prompts": "Design baseline prompts for the vLLM to process the entire video directly without decomposition. Use state-of-the-art models (e.g., GPT-4 vision API or mPLUG-Owl) with prompts like: Baseline Prompt: "Summarize the events of the entire video. Focus on describing all key actions and objects. Output expectations for these prompts will be used for comparison later.",
                \item\sethlcolor{lightgreen}\hl{"Step 4: Construct Temporal-Decomposition Prompts": "Design decomposition-specific prompts for:(1) Segment Summarization:- Example Prompt: "For Segment [X time interval], summarize the main events. Be specific about what happens, who or what is involved, and their actions."(2) Segment Alignment and Reasoning:- Example Prompt: "Combine the key events from Segment X and Segment Y. Are the events causally related? If yes, explain how. If not, infer temporal dependencies like 2018 before 2019 or 2018 after 2019."(3) Full Video Reconstruction:- Example Prompt: "Using the summaries of all segments, reconstruct the full narrative, maintaining chronological order. Highlight any interactions between entities across the segments.",}
                \item "Step 5: Select Models": "We test GPT-4 Vision API as the main vLLM model along with comparison models such as mPLUG-Owl and VideoChat-Embed. For open-source options, use LLaVA, mPLUG, and MiniGPT-4 (if the videos can be converted into image-frame-based inputs for these models).",
                \item "Step 6: Generate Results": "Feed the video clips into the models using both baseline prompts and TDP prompts. Record outputs for all test videos. Extract segment-level summaries, alignment narratives, and final video summaries.",
                \item  "Step 7: Evaluate Results": "Use the following evaluation metrics:1. **Causal Inference Accuracy (Ego4D)**: Measure how accurately the model predicts causal relationships in video events.2. **Temporal Alignment Consistency (VidSitu)**: Quantify correctness in event ordering and alignment scores provided in VidSitu annotations.3. **Long-term Memory Probing**: Test the model's ability to recall events from early video segments, ensuring no significant context loss.4. **Human Judgments (Optional)**: Inspect outputs qualitatively for coherence and alignment using comparison methods where automated metrics are limited.",
                \item  "Step 8: Analyze and Compare Baselines vs TDP": "Perform comparative analysis to show the advantages of TDP over baseline methods. Focus on:1. Ability to preserve context from start to end of long videos.2. Consistency in causal reasoning across multiple segments.3. Reductions in memory issues and hallucinations.",
                \item "Step 9: Perform Ablations": "Test the performance of TDP by selectively removing one of the key steps (e.g., omit iterative refinement or alignment reasoning prompts) to evaluate which components are most critical."
        \end{itemize}
   \end{itemize}
\end{itemize}

\vspace{2\baselineskip}

\textbf{AI-Scientist:}
\begin{itemize}
\item \textbf{Motivation:} Existing vLLMs struggle with long-form video understanding due to memory constraints, inefficient temporal reasoning, and short temporal certificates in datasets. VideoChat's VLTF and Video-LLaMA's Q-formers partially address cross-modality alignment but falter with long-term dependencies. EgoSchema emphasizes temporal hardness but lacks a memory-aware approach. SEED-Bench provides generative comprehension evaluation but is limited in fine-grained temporal understanding. \sethlcolor{lightgreen}\hl{This work aims to introduce a memory-augmented framework to efficiently manage long-term temporal dependencies and improve temporal reasoning capabilities in vLLMs.}
\item \textbf{Experimental plan:} 
   \begin{itemize}
   \item \textbf{Experiment step:}
        \begin{itemize}
               \item "Step1 Develop a memory-augmented vLLM architecture.":  "\sethlcolor{lightgreen}\hl{Introduce a memory module (e.g., transformer-based memory or recurrent memory) within the VLTF/Q-former architecture to retain temporal information across video segments.}",
                            "Implement a memory update mechanism that compresses past video frames and dialogues into high-dimensional memory embeddings, preserving critical spatiotemporal information.",
                            "\sethlcolor{lightgreen}\hl{Modify the attention mechanism to retrieve relevant information from the memory module for reasoning over long-term dependencies.}"
                \item  "Step2 Temporal-aware sampling and dataset preparation.":  "Modify existing video datasets (e.g., WebVid-10M, SEED-Bench) by introducing a **Memory Temporal Certificate** (MTC) metric, which evaluates how well models utilize retained temporal knowledge.",
                            "Split long videos into chunks, ensuring temporal overlap between consecutive chunks, and integrate memory-aware instructions during video question-answer dialogues.",
                            "Construct tasks involving incremental temporal inference (e.g., predicting future actions based on past segments)."
                \item  "Step3 Model training and evaluation.": "Pre-train the memory-augmented vLLM on WebVid-10M and HowTo100M datasets using video-to-text generation tasks, incorporating the memory module during training.",
                            "Fine-tune the model on SEED-Bench and evaluate tasks requiring temporal reasoning through MTC and other established metrics like temporal retention accuracy and incremental response accuracy.",
                            "Compare performance against baseline models (e.g., VideoChat, Video-LLaMA) to quantify improvements in long-term video understanding."
        \end{itemize}
   \end{itemize}
\end{itemize}

\vspace{2\baselineskip}
\textbf{VIRSC:}
\begin{itemize}
\item \textbf{Motivation:} This paper proposes a next-generation system for long video understanding in vLLMs called **Schema-Driven Causal vLLM (SDC-vLLM)**. The system introduces a novel approach to explicitly decompose long videos into hierarchical event schemas and build causal reasoning chains across these schemas. The core innovation lies in three key components: (1) **Event Schema Abstraction Module (ESAM)**, a graph-based mechanism that segments long videos into meaningful, high-level event schemas by clustering spatiotemporal features and aligning them with semantic event patterns; (2) \sethlcolor{lightgreen}\hl{**Causal Chain Modeling (CCM)**, a reinforcement learning-based module that dynamically learns the causal dependencies between events, capturing both local and long-range cause-effect relationships across the video's timeline;} and (3) **Event-Driven Adaptive Memory (EDAM)**, \sethlcolor{lightgreen}\hl{an episodic memory system optimized to store and retrieve event representations} and causal links, reducing redundancy and focusing on high-level reasoning. The system integrates these components with hierarchical modeling and adaptive compression to achieve efficient and scalable understanding of long videos. By emphasizing schema-driven reasoning and causality, SDC-vLLM fills a critical gap in the vLLM landscape, enabling advanced tasks such as causal inference, event prediction, and multi-step reasoning in video-based question answering.
\item \textbf{Experimental plan:} 
   \begin{itemize}
   \item \textbf{Experiment step:}
        \begin{itemize}
               \item  "Step1: Dataset Construction": "Construct a novel dataset, **CausalEvents-10K**, specifically tailored to long video causal reasoning. Extend existing datasets (e.g., YouTube-8M, ActivityNet, Ego4D) by annotating long videos with event schema metadata and causal links between events. Use GPT-4 to assist in generating causal event annotations and multi-step reasoning questions. Human curation ensures high-quality, diverse, and temporally challenging samples spanning $>=$10-minute durations. Each video is paired with hierarchical event schema graphs and causal annotations, emphasizing abstract, high-level structures.",
                \item    " "Step2: Model Design": 
                \begin{itemize}
                    \item  "Event Schema Abstraction Module (ESAM)": "Develop a graph-based clustering algorithm that processes sliding window video embeddings to identify recurring high-level events. Each event schema is represented as a graph node, with temporal relationships forming edges. Pre-trained semantic embeddings align video segments with known event patterns, providing interpretable schema representations.",
                    \item "Causal Chain Modeling (CCM)": "Train a reinforcement learning agent to map causal relationships between events. The agent learns to traverse the event schema graph, identifying local and global causal dependencies across the timeline. Supervised pretraining on causal datasets and policy gradient fine-tuning ensure robust cause-effect reasoning.",
                    \item "Event-Driven Adaptive Memory (EDAM)": "Design an episodic memory bank optimized for event schema storage. The memory system dynamically ranks and stores key events and their causal links, removing irrelevant or redundant details. It integrates hierarchical aggregation, compressing lower-level spatiotemporal features into schema-centric memory representations.",
                    \item "Integration with LLM": "Incorporate the schema and causal representations into an LLM such as GPT-4, enabling multi-round reasoning. The LLM is fine-tuned to process event schemas and causal graphs, enabling it to answer multi-step questions requiring deep causal inference and event prediction."
                \end{itemize}
                \item  "Step3: Training":
                \begin{itemize}
                    \item "Stage1: Pretraining": "Pretrain the video encoder, ESAM, and CCM on large-scale video-text datasets, ensuring alignment between spatiotemporal features, event segmentation, and causal reasoning embeddings. Use datasets like Ego4D and YouTube-8M as foundational corpora.",
                    \item "Stage2: Instruction Fine-Tuning": "Fine-tune the entire system on CausalEvents-10K, focusing on multi-step reasoning tasks, causal inference, and event prediction. Use diverse instruction types such as 'What caused Event X?' or 'What is the expected outcome of Event Y?'
                    \item "Stage3: Memory Optimization": "Optimize EDAM using reinforcement learning to ensure efficient memory storage and retrieval. The reward signal prioritizes the retention of causally important events while discarding irrelevant details, improving long-term reasoning efficiency."
                \end{itemize}
                \item   "Step4: Evaluation":
                \begin{itemize}
                    \item "Metrics": "Evaluate using benchmarks such as ActivityNet-QA, Charades, and the new CausalEvents-10K. Key metrics include causal inference accuracy, temporal reasoning F1-score, and multi-step QA success rate. Introduce novel metrics like 'Causal Chain Completion' to assess the system's ability to reconstruct cause-effect relationships in unseen videos.",
                    \item "Baselines": "Compare SDC-vLLM against existing models like HMA-vLLM, EgoSchema-based benchmarks, and VideoChat. Assess performance on tasks such as causal inference, event prediction, and schema-based QA.",
                    \item  "Ablation Studies": "Conduct extensive ablations to analyze the contributions of ESAM, CCM, EDAM, and the hierarchical modeling pipeline. Test performance with and without causal annotations to evaluate the impact of explicit causality modeling."
                \end{itemize}       
        \end{itemize}
   \end{itemize}
\end{itemize}

\vspace{2\baselineskip}
\textbf{SCIPIP}
\begin{itemize}
\item \textbf{Motivation:}  **Hierarchical Temporal Graph Memory with Multimodal Integration**- **Details**:  Extend the hierarchical memory framework by constructing a **Hierarchical Temporal Graph (HTG)** where temporal segments (e.g., frames, shots, scenes) are represented as nodes across multiple abstraction levels. Graph Neural Networks (GNNs) encode the intra-level relationships between nodes (e.g., action similarity in frames, semantic coherence across scenes) while inter-level attention mechanisms facilitate information flow across levels of hierarchy. Each node supports **multimodal embeddings** (video frames, audio waveforms, subtitles) to unify representations across modalities.  \sethlcolor{lightgreen}\hl{Temporal coherence is further enhanced using **adaptive graph updates**, where low-importance segments (identified through semantic scoring) are compressed into merged nodes, and high-importance segments are retained in fine-grained form.} Knowledge is propagated across the graph hierarchy using memory-enhancing mechanisms such as Transformer-based attention layers. This representation captures both granular details (at low levels) and abstract/event-level summaries (at high levels), making it well-suited for long-form video reasoning.
\end{itemize}
\end{tcolorbox}
\clearpage

\section{Case 4}
\label{appendix:Case_4}
\begin{tcolorbox}[breakable,enhanced,width=\textwidth]
\textbf{Target paper:} 
R-tuning: Instructing large language models to say ‘I don’t know’~\cite{zhang2024r}

\textbf{Target motivation:} 

LLMs have demonstrated remarkable performance across various tasks, but they often generate non-existent facts, a phenomenon known as hallucination. Current instruction tuning methods force models to complete sentences regardless of their knowledge, leading to fabricated responses when questions are beyond their knowledge. This paper addresses the gap between instruction tuning data and parametric knowledge, proposing a method to teach models to refuse unknown questions, thereby mitigating hallucination and improving model reliability.

\textbf{Summary of target experiment:} 

The paper introduces Refusal-Aware Instruction Tuning (R-Tuning), a two-step approach: (1) identifying uncertain questions by measuring the knowledge gap between instruction tuning data and parametric knowledge, and (2) constructing refusal-aware data by appending uncertainty expressions to uncertain questions, then fine-tuning the model on this data.

\textbf{Designs of target experiment:} 
\begin{itemize}
\item \textbf{Design 1:} 
   \begin{itemize}
   \item \textbf{Design name:}  Refusal-Aware Data Identification
   \item \textbf{Description of design name:} The first step involves identifying uncertain questions by comparing the model's predictions with ground-truth labels. Questions are split into certain (D1) and uncertain (D0) sets based on whether the model's prediction matches the label. This step uses a supervised strategy to recognize knowledge gaps between the model's parametric knowledge and the training data.
   \end{itemize}
\item \textbf{Design 2:} 
   \begin{itemize}
   \item \textbf{Design name:}  Refusal-Aware Data Construction
   \item \textbf{Description of design name:} The second step constructs refusal-aware data by appending uncertainty expressions to uncertain questions and certainty expressions to certain questions. This step uses a padding method to teach the model to express uncertainty when answering uncertain questions and confidence when answering certain questions.
   \end{itemize}
\end{itemize}

\textbf{Topic:} 
The topic of this paper is mitigating hallucination in large language models by teaching them to refuse unknown questions.

\textbf{Inspiration papers:} 
\begin{itemize}
\item Selfcheckgpt: Zero-resource black-box hallucination detection for generative large language models~\cite{manakul2023selfcheckgpt}
\item On faithfulness and factuality in abstractive summarization~\cite{maynez2020faithfulness}
\item Measuring massive multitask language understanding~\cite{hendrycks2020measuring}
\item On calibration of modern neural networks~\cite{guo2017calibration}
\item Llama: Open and efficient foundation language models~\cite{touvron2023llama}
\end{itemize}

\vspace{2\baselineskip}

\textbf{AI-Researcher:}
\begin{itemize}
\item \textbf{Motivation:} Prior methods like SelfCheckGPT and factual consistency scoring aim to detect hallucinations but still operate with the assumption that LLMs must provide an answer to every query. These approaches do not teach models to refuse answers when faced with topics outside their expertise. Our proposed method, Knowledge-Boundary Prompting, addresses this limitation by explicitly instructing LLMs to identify their knowledge boundaries and refuse to answer when uncertain. \sethlcolor{lightgreen}\hl{By focusing on refusal-based reasoning and providing clear refusal language, we aim to reduce hallucinations and improve user trust.} This approach leverages LLMs' responsiveness to carefully designed prompts, avoiding the need for additional supervision or human intervention in real-time interactions.
\item \textbf{Experimental plan:} 
   \begin{itemize}
   \item \textbf{Proposed Method:} \sethlcolor{lightgreen}\hl{Knowledge-Boundary Prompting introduces a multi-step prompting framework that teaches LLMs to refuse queries when uncertain. The method involves:1. **Self-Evaluation:** The LLM is prompted to assess whether the query falls within the scope of its training knowledge by asking, "Have you encountered this topic during earlier training phases?"}2. **Refusal-Based Reasoning:** If uncertain, the LLM generates reasoning to justify refusal, for example, "This query requires external information not present in the training data."3. **Refusal Language Formulation:** The LLM is prompted to provide explicit refusal responses, such as "I cannot confidently answer this."By breaking down the refusal process into these structured steps, the proposed method leverages the LLM's existing capabilities for reasoning and natural language expression, steering it toward safer behavior in scenarios involving uncertainty.
   \item \textbf{Experiment step:}
        \begin{itemize}
               \item "Step 1: Gather Datasets": "\sethlcolor{lightgreen}\hl{We evaluate the efficacy of Knowledge-Boundary Prompting using factual QA datasets that include both answerable and unanswerable questions}: 1. WebQuestions (for knowledge-intensive questions). 2. Natural Questions (NQ) (for real-world queries with varying complexity). 3. BoolQ (for binary yes/no questions). For unanswerable queries, we use the *Unanswerable Questions* dataset (e.g., from SQuAD 2.0 or custom-generated unanswerable questions)."
                \item  "Step 2: Construct Prompts": "We design the following prompts to evaluate the baseline and the proposed Knowledge-Boundary Prompting method:**Baseline Prompts:**- Direct Prompting: Provide the query directly to the LLM, e.g., "Who was President of the United States in 1800?"**Proposed Method Prompts:**1. **Self-Evaluation Prompt:**   Input: "[Query]. Have you encountered this topic before during earlier training phases? Respond 'yes' if this is within your knowledge, 'no' if unknown, or 'unsure' if not fully certain."2. **Refusal-Based Reasoning Prompt:** (Used when self-evaluation indicates "no" or "unsure")   Input: "Based on your assessment, explain why you cannot answer the query. For example: 'This query requires external information not in my training data.'"3. **Refusal Language Prompt:** (Used after refusal-based reasoning)   Input: "Given your uncertainty, respond with an explicit refusal to answer, such as: 'I cannot confidently answer this. Please consult a domain expert or source like Wikipedia.'"We also design few-shot prompting experiments by demonstrating refusals over similar questions in the prompt context.",
                \item"Step 3: Select Models": "Evaluate the method on the following LLMs:- GPT-3.5 (gpt-3.5-turbo, OpenAI API).- GPT-4 (gpt-4, OpenAI API).- Claude 2 (Anthropic API).- Open-source large models such as LLaMA-2-70B-chat.",
                \item"Step 4: Experimental Setup and Execution":
                \begin{itemize}
                    \item "1. Query Dataset Sampling": "Select 500 answerable and 500 unanswerable queries from the datasets, ensuring balanced representation. Specifically, unanswerable queries should cover a range of uncertainty (e.g., ambiguous questions, factual errors, under-specified queries).",
                    \item "2. Baseline Benchmarks": "Run baseline direct prompting on all queries and record the outputs for comparison. Metrics include:- **Hallucination Rate (HR):** Percentage of unanswerable queries where the LLM provides an incorrect answer instead of refusing.- **Correct Answer Rate (CAR):** Percentage of answerable queries where the LLM provides the correct answer.",
                    \item "3. Apply Knowledge-Boundary Prompting": "Run the proposed multi-step prompting method on the same set of queries and record the outputs for:   a. \sethlcolor{lightgreen}\hl{**Self-Evaluation Predictions:** Record whether the LLM classified each query as 'known,' 'unknown,' or 'unsure.'   b. **Refusal Reasoning:** Generate reasons for refusals and validate their correctness through manual inspection.   c. **Refusal Responses:** Evaluate if the language of the refusal explicitly conveys uncertainty (manually or via a scoring script for refusal phrases)."},
                    \item "4. Metrics Collection": "Compare results across metrics: (i) **Refusal Accuracy (RA):** Percentage of unanswerable queries correctly classified as unknown and refused. (ii) **Confidence Alignment (CA):** Percentage of answerable queries classified as 'known' or 'unsure' with correct answers.(iii) Compare HR, CAR, and RA between baseline and proposed method."
                \end{itemize}
                \item "Step 5: Analyze Results": "Conduct statistical analyses to compare baseline and proposed methods. Key analyses include: - Change in HR between baseline and Knowledge-Boundary Prompting. - Consistency of refusal accuracy (RA) across datasets. - Performance variations across GPT-3.5, GPT-4, Claude 2, and LLaMA-2-70B-chat."
        \end{itemize}
   \end{itemize}
\end{itemize}

\vspace{2\baselineskip}

\textbf{AI-Scientist:}
\begin{itemize}
\item \textbf{Motivation:} \sethlcolor{lightgreen}\hl{Large Language Models (LLMs) are prone to hallucinating when presented with queries that are unverifiable or beyond their training data.} While prior research has explored hallucination detection methods (e.g., SelfCheckGPT) and calibration techniques (e.g., temperature scaling), these methods focus on post-hoc evaluations rather than proactive behavioral training. \sethlcolor{lightgreen}\hl{This paper aims to address this gap by introducing a framework that teaches LLMs to refuse answering unverifiable questions while maintaining performance on factual queries.} The inclusion of dynamic confidence thresholds and user-centric evaluation metrics ensures scalability and generalization of the framework across domains.
\item \textbf{Experimental plan:} 
   \begin{itemize}
   \item \textbf{Experiment step:}
        \begin{itemize}
               \item "Step1": "Dataset Creation - \sethlcolor{lightgreen}\hl{Generate queries with verifiable and unverifiable content using existing datasets like WikiBio and XSUM. Label queries as 'factual,' 'hallucinated,' or 'unknown' based on human annotation. Additionally, introduce domain-specific variations to ensure robustness across multiple contexts.}",
                \item  "Step2": "Model Modification - Implement temperature scaling to calibrate the model's confidence on output probabilities. Introduce a dynamic thresholding mechanism that adjusts the refusal confidence threshold based on query complexity and domain.",
                \item "Step3": "Training - Train the LLM using a multitask objective: (i) factual queries are answered normally, (ii) hallucinated content is flagged, and (iii) unknown queries result in a refusal response such as 'I cannot answer this question.' Incorporate reinforcement learning to reward appropriate refusal behavior based on dynamic thresholds.",
            \item "Step4": "Evaluation - Evaluate the model on a diverse test set of queries covering multiple domains (e.g., medicine, law, psychology). Metrics include accuracy for factual queries, precision for hallucination detection, refusal rate for unknown questions, and user satisfaction scores for refusal behavior measured via simulated evaluations.",
            \item "Step5": "Ablation Study - Test the importance of dynamic thresholding, temperature scaling, and multitask training by removing each component and measuring performance drops.",
            \item "Step6": "Error Analysis - Analyze failure cases to identify patterns in the model's refusal behavior, particularly focusing on borderline queries where confidence is ambiguous."
        \end{itemize}
   \end{itemize}
\end{itemize}

\vspace{2\baselineskip}
\textbf{VIRSC:}
\begin{itemize}
\item \textbf{Motivation:} Building on the reactive retrieval mechanism for mitigating hallucination in LLMs, this idea introduces a 'Proactive Knowledge Gap Identification and Dynamic Retrieval Framework' (PKG-DRF) to augment LLMs with anticipatory epistemic capabilities. The system operates in two layers: \sethlcolor{lightgreen}\hl{(1) A proactive knowledge gap detection module that evaluates the model's internal knowledge base against domain-specific benchmarks and real-time updates. This module identifies potential gaps before deployment by self-assessing the model's knowledge across key domains and noting areas that require supplemental information;} and (2) A dynamic retrieval mechanism that periodically queries trusted external databases and updates the model's external knowledge graph in real time. Additionally, the system incorporates a domain-adaptive hallucination prevention mechanism, enabling it to fine-tune retrieval and response generation strategies based on domain-specific needs (e.g., stricter factuality thresholds in healthcare, balanced reasoning in law). Unlike traditional reactive systems that address hallucination only when unknown queries arise, this framework proactively increases the model's preparedness and adaptability across dynamic use cases. This approach not only mitigates hallucination but also enhances the model's contextual awareness, factuality, and user trust in real-world, high-stakes applications.
\item \textbf{Experimental plan:} 
   \begin{itemize}
   \item \textbf{Experiment step:}
        \begin{itemize}
               \item  "Step1": "\sethlcolor{lightgreen}\hl{Develop a multi-domain evaluation and self-assessment dataset to train the proactive knowledge gap detection module.} This dataset should include structured benchmarks for domains such as healthcare, law, and education, as well as real-time updates like recent medical discoveries or new legal precedents. Use human experts to annotate gaps in the model's performance for these tasks, creating a training signal for the proactive evaluation mechanism.",
                \item   "Step2": "Fine-tune the proactive knowledge gap detection module on the multi-domain dataset. Train the model to assign 'knowledge completeness scores' across domains, identifying areas where its internal knowledge is insufficient or outdated. Integrate this module into the model's pretraining and fine-tuning pipeline to dynamically assess and flag knowledge gaps during training.",
                \item  "Step3": "Develop a dynamic retrieval mechanism designed to periodically access and integrate information from trusted, domain-specific knowledge bases. This mechanism should update a lightweight external knowledge graph that the LLM can reference during inference. For example, medical information can be retrieved from PubMed, legal data from government portals, and educational content from standardized academic databases.",
                \item "Step4": "Combine the proactive knowledge gap detection module with the retrieval mechanism to enable real-time gap assessment and retrieval during both fine-tuning and deployment. For known knowledge gaps, retrieve and integrate the missing information; for ambiguous or speculative queries, the model should either refuse to answer or provide a transparent response citing its limitations or the retrieved sources.",
                \item "Step5": "Design a domain-adaptive hallucination prevention mechanism that adjusts retrieval precision and response construction based on the domain. For instance, in high-stakes applications like healthcare, prioritize factual accuracy and verifiability by using stricter entailment techniques. In less critical contexts like general education, balance factuality with creativity.",
                \item "Step6": "Evaluate the system on both static and dynamic benchmarks. Use established benchmarks such as TruthfulQA and LAMA for static evaluation, and introduce dynamic benchmarks that simulate real-world scenarios with evolving knowledge requirements (e.g., new medical guidelines, legal rulings, or academic discoveries). Metrics should include hallucination rates, retrieval relevance, knowledge gap coverage, and response transparency.",
                \item "Step7": "Test the model in real-world, high-stakes domains with end-users and domain experts. Analyze qualitative and quantitative metrics such as user trust scores, error mitigation rates, and domain-specific performance. Identify edge cases where proactive retrieval or domain-adaptive mechanisms fail, and refine the system iteratively to address these gaps."
        \end{itemize}
   \end{itemize}
\end{itemize}

\vspace{2\baselineskip}
\textbf{SCIPIP}
\begin{itemize}
\item \textbf{Motivation:}   Multi-Modal Dynamic Confidence Calibration and Refusal System (MM-DCCRS)**  - **Details**:  Merge **Dynamic Multi-Modal Confidence and Uncertainty Alignment (DMC-UA)** and elements of **Metacognition Module with Contradiction Injection (MCM-CI)** into a streamlined yet powerful framework. This approach dynamically calibrates confidence levels across multiple modalities (e.g., text, retrieved knowledge, images) using a probabilistic scoring system. \sethlcolor{lightgreen}\hl{The model evaluates internal consistency by cross-referencing confidence across modalities. Divergence in confidence metrics or detection of ambiguous data triggers the refusal mechanism.} The calibration system is further enhanced by training with contradiction injection, where ambiguous or contradictory queries are deliberately fed into the model to strengthen its ability to detect and respond appropriately to uncertainty.
\end{itemize}
\end{tcolorbox}
\clearpage

\section{Prompt for summary target paper.}

\label{appendix:Summary_Paper}
\begin{tcolorbox}[breakable,enhanced,width=\textwidth]

\textbf{System Prompt:}

You are an AI assistant tasked with summarizing research papers in a structured and clear manner.

\vspace{0.3em}  
\textbf{User Instructions:}  
Please summarize the following research paper by providing the following details in JSON format: 
\begin{itemize}
    \item \textbf{Topic:} Identify the key concepts, research questions, or objectives discussed in the paper. Summarize the main topic in one or two sentences, ensuring it captures the essence of the paper. Avoid including unnecessary details or examples.
  
    \item \textbf{Motivation:} Provide an explanation of the current state of the field. What are the key achievements in this area of research, and what are the limitations or open challenges that this paper addresses? Describe why this research is important and how it aims to contribute to advancing the field.
    
    \item \textbf{Method:} Explain the methodology used in the paper. In particular, describe the specific designs or approaches the authors implemented to address the limitations or gaps identified in the motivation. Provide a summary of the targeted designs, followed by a detailed explanation of each design individually. 
\begin{itemize}
\item \textbf{Summary:} Give an overview of the key innovations or design choices made to overcome existing limitations in the field.
\item \textbf{Detailed designs:} For each targeted design, provide a thorough explanation. Discuss innovations in model architecture, algorithms, data processing techniques, or training strategies. Please anonymize the names of the methods in the descriptions.
\item \textbf{Problems solved:} List the specific problems that each design addresses, separately.
\item \textbf{Datasets:} Mention the datasets used for training and testing the model, including any unique characteristics or challenges they present.
\item \textbf{Metrics:} Specify the evaluation metrics used to assess the model's performance, such as accuracy, precision, recall, F1 score, etc.
\end{itemize} 
\end{itemize}  

Ensure that the output adheres to the following requirements:  
\begin{itemize}
    \item \textbf{Provide clear and concise explanations.}  
    \item \textbf{Summarize the content in a structured, easy-to-read format.}  
    \item \textbf{For the "method" section, ensure that:}
    \begin{itemize}
        \item \textbf{Targeted designs:} The summary should provide an overview of the key innovations or strategies. 
        
        \item \textbf{Individual designs:} Determine what small designs make up the overall framework, break down the whole framework into individual small design and use the orignal sentece from the paper to explain them in detail separately, such as the detail architectural changes, novel algorithms, or new techniques. Anonymize the names of methods and techniques by describing them in a general sense, avoiding any specific names.
  
        \item \textbf{Problems solved:} List the specific problems each design addresses separately
        \item \textbf{Datasets:} Mention the datasets used for training and testing the model, including any unique characteristics or challenges they present.
        \item \textbf{Metrics:} Specify the evaluation metrics used to assess the model's performance, such as accuracy, precision, recall, F1 score, etc.
        
        \end{itemize}
\end{itemize}
Use the following JSON structure for the output:

\begin{minted}[breaklines, breakanywhere=true, breaksymbolleft={},
    breaksymbolright={}]{python}
{
  "topic": "The main research object and scope of the study",
  "motivation": "Current state of the field, achievements, and limitations addressed by this study",
  "method": {
    "targeted_designs_summary": "A high-level summary of the designs or innovations made to address limitations",
    "targeted_designs_details": [
      {
        "design_name": "Name of the design",
        "description": "Detailed explanation of this design, including its purpose, how it addresses limitations, and any novel aspects (anonymized)",
        "problems_solved": "Problem that this design solve"
      },
      {
        "design_name": "Name of another design",
        "description": "Detailed explanation of this design (anonymized)",
        "problems_solved": "Problem that this design solve"
      }
    ],
    "datasets": "Datasets used in the experiments",
    "metrics": "Evaluation metrics used to assess the effectiveness of the approach"
  }
}
\end{minted}

Here is the provided paper:

[Paper Content]
\end{tcolorbox}

\clearpage

\section{Prompt for find most cited paper.}

\label{appendix:Find_Cited_Paper}
\begin{tcolorbox}[breakable,enhanced,width=\textwidth]

\textbf{System Prompt:} 

You are an academic assistant tasked with analyzing a research paper and identifying the top most-cited references within it. For each reference, provide the title, the cited number, and the sections where the reference is cited, along with the cited number in each section.

\vspace{0.3em}  
\textbf{User Instructions:}  

Please follow the following steps for analysis:
\begin{itemize}
\item Read the entire thesis carefully and count the number of citations for each reference.
\item Identify the top ten references with the highest number of citations.
\item For each selected reference, determine its title, the total number of citations, as well as the chapters in which it is cited and the number of citations in each chapter.
\item Output the results in the following JSON structure in descending order of the number of citations: 
\end{itemize}

Use the following JSON structure for the output:

\begin{minted}[breaklines, breakanywhere=true, breaksymbolleft={},
    breaksymbolright={}]{python}
{
        "top_references": [
            {
                "rank": 1,
                "title": "A Comprehensive Study on Machine Learning Techniques",
                "cited_number": 5,
                "sections": [
                    {"name": "Introduction", "cited_number": 3},
                    {"name": "Literature Review", "cited_number": 2}
                ]
            },
            {
                "rank": 2,
                "title": "Deep Learning for Natural Language Processing",
                "cited_number": 5,
                "sections": [
                    {"name": "Methodology", "cited_number": 3},
                    {"name": "Results", "cited_number": 2}
                ]
            },
            {
                "rank": 3,
                "title": "Neural Networks and Their Applications",
                "cited_number": 4,
                "sections": [
                    {"name": "Literature Review", "cited_number": 2},
                    {"name": "Discussion", "cited_number": 2}
                ]
            },
            {
                "rank": 4,
                "title": "Understanding AI Algorithms in Modern Computing",
                "cited_number": 4,
                "sections": [
                    {"name": "Introduction", "cited_number": 3},
                    {"name": "Conclusion", "cited_number": 1}
                ]
            },
            {
                "rank": 5,
                "title": "Advances in Supervised Learning",
                "cited_number": 3,
                "sections": [
                    {"name": "Methodology", "cited_number": 3}
                ]
            },
            {
                "rank": 6,
                "title": "Applications of AI in Healthcare",
                "cited_number": 3,
                "sections": [
                    {"name": "Literature Review", "cited_number": 2},
                    {"name": "Discussion", "cited_number": 1}
                ]
            },
            {
                "rank": 7,
                "title": "A Survey of Reinforcement Learning Techniques",
                "cited_number": 2,
                "sections": [
                    {"name": "Introduction", "cited_number": 2}
                ]
            },
            {
                "rank": 8,
                "title": "The Evolution of Neural Network Models",
                "cited_number": 2,
                "sections": [
                    {"name": "Methodology", "cited_number": 2}
                ]
            },
            {
                "rank": 9,
                "title": "Optimization Methods in Machine Learning",
                "cited_number": 2,
                "sections": [
                    {"name": "Literature Review", "cited_number": 1},
                    {"name": "Results", "cited_number": 1}
                ]
            },
            {
                "rank": 10,
                "title": "Data Mining and Its Challenges",
                "cited_number": 1,
                "sections": [
                    {"name": "Conclusion", "cited_number": 1}
                ]
            }
        ]
    }
\end{minted}

Here is the provided paper:

[Paper Content]
\end{tcolorbox}

\clearpage
\section{Prompt for clean topic.}

\label{appendix:clean_topic}
\begin{tcolorbox}[breakable,enhanced,width=\textwidth]

\textbf{System Prompt:} 

You are an expert in academic writing and text analysis. Your task is to evaluate whether a given topic statement is concise, specific, and focused on the core theme. The topic should avoid unnecessary details or examples that don’t directly contribute to the core concept. Your goal is to remove unnecessary content, leaving just the essential theme.

Provide your output in JSON format, following the instructions precisely.

\vspace{0.3em}  
\textbf{User Instructions:}  
\begin{itemize}
\item Carefully read the provided topic statement and motvaiton.
\item Assess if the statement contains unnecessary details, such as specific methods, examples, or tangential content. Focus on identifying any overly specific explanations that don't contribute to the core subject.
\item If unnecessary details are found, suggest a revised version of the topic that keeps only the main subject, removing extraneous content. The revised version should follow the format: "The topic of this paper is [revised topic].
\item If no unnecessary details are found, confirm that the topic is concise and specific.
\item Format your output as a JSON object with the following keys:
\begin{itemize}
   \item original\_topic: The original topic statement provided.
   \item contains\_unnecessary\_details: A boolean value (true or false) indicating whether unnecessary details or examples are present.
   \item revised\_topic: If unnecessary details are found, provide a revised version of the topic statement in the format "The topic of this paper is [revised topic]." If no unnecessary details are found, set this to null.
\end{itemize}
\end{itemize}

Example Input:
\begin{minted}[breaklines, breakanywhere=true, breaksymbolleft={},
    breaksymbolright={}]{python}
"The paper introduces a novel Part Re-projection Distance Loss (PRDL) for 3D face reconstruction, leveraging facial part segmentation to improve alignment and reconstruction accuracy, especially for extreme expressions."
\end{minted}
Example Output:

\begin{minted}[breaklines, breakanywhere=true, breaksymbolleft={},
    breaksymbolright={}]{python}
{
"original_topic": "The paper introduces a novel Part Re-projection Distance Loss (PRDL) for 3D face reconstruction, leveraging facial part segmentation to improve alignment and reconstruction accuracy, especially for extreme expressions.",
"contains_unnecessary_details": True,
"revised_topic": "The topic of this paper is 3D face reconstruction."
}
\end{minted}

Here is a motivation for reference:

[Motivation]

Now, evaluate the following topic statement:

[Topic]

\end{tcolorbox}

\clearpage
\section{Prompt for split topic.}

\label{appendix:split_topic}
\begin{tcolorbox}[breakable,enhanced,width=\textwidth]

\textbf{System Prompt:}

You are an assistant specializing in extracting key, relevant keywords from a provided topic. Your goal is to identify specific keywords that directly represent the corefocus of the text. These keywords should be clear, precise, and directly linked to the topic's content. Avoid using broad, vague, or overly general terms.

\vspace{0.3em}  
\textbf{User Instructions:}  
\begin{itemize}
\item Carefully read the provided topic to fully understand its core focus.
\item Extract specific, important keywords that are closely tied to the core themes, concepts, or findings of the topic. These keywords should be essential to understanding the subject matter.
\item Rank the extracted keywords by relevance, starting with the most important.
\item For each keyword, provide a concise explanation that justifies its ranking and details how it connects to the main focus of the topic.
\item The output should be in JSON format with each keyword and its explanation as a dictionary object.
\end{itemize}

Input:

Topic: [Topic]

Output:
\begin{itemize}
\item "rank": The position of the keyword based on its relevance, with 1 being the most important.
\item "keyword": The specific keyword that represents a key concept related to the topic.
\item "explanation": A short justification for why this keyword is central to the topic.
\end{itemize}

Output Format:

The output should be in JSON format with the following structure:

\begin{minted}[breaklines, breakanywhere=true, breaksymbolleft={},
    breaksymbolright={}]{python}
 {
    "keywords": [
        {
            "rank": 1,
            "keyword": "Keyword 1",
            "explanation": "Explanation of why this keyword is the most relevant."
        },
        {
            "rank": 2,
            "keyword": "Keyword 2",
            "explanation": "Explanation of why this keyword is the second relevant."
        },
        ...
    ]
}
\end{minted}
\end{tcolorbox}

\clearpage
\section{Prompt for get content of reference paper.}

\label{appendix:content_reference}
\begin{tcolorbox}[breakable,enhanced,width=\textwidth]

\textbf{System Prompt:}

You are a scientific research expert, tasked with extracting and summarizing information from provided paper content relevant to the topic: {topic}. Your deliverables will include pertinent references, extracted entities, a detailed summary, and the experimental design.

\vspace{0.3em}  
\textbf{User Instructions:}  

The topic you are studying is: [topic]. (Ensure that the references are pertinent to this topic.)

Extraction Requirements:

Entities
\begin{itemize}
\item Identify unique entities mentioned in the paper, such as model names, datasets, metrics, and specialized terminology.
\item Format the entities with a name followed by a brief description.
\item Ensure all entities are relevant to the specified topic ([topic]).
\end{itemize}

Summary Idea:
\begin{itemize}
\item Background: Elaborate on the task's context and previous work, outlining the starting point of this paper.
\item Novelty: Describe the main innovations and contributions of this paper in comparison to prior work.
\item Contribution: Explain the primary methods used, detailing the theory and functions of each core component.
\item Detail Reason: Provide a thorough explanation of why the chosen methods are effective, including implementation details for further research.
\item Limitation: Discuss current shortcomings of the approach.
\end{itemize}

Relevance Criteria:
\begin{itemize}
\item Method Relevance: References must directly correlate with the paper's methodology, indicating improvements or modifications.
\item Task Relevance: References should address the same task, even if methods differ, better have the same topic {topic}.
\item Baseline Relevance: References should serve as baselines for the methods discussed in the paper.
\item Output Format: Provide references without author names or publication years, formatted as titles only.
\item Specific paper titles will be placed between <References></References>. Based on the precise citation location and the corresponding ref\_id in the paper, you need to infer the specific title of your output relevant references.
\end{itemize}
The paper content is as follows: 
[paper content]

Please provide the entities, summary idea, experimental design, and the three most relevant references (Sort by relevance, with priority given to new ones with the same level of relevance, do not reference the original paper.) based on the paper's content.
Note: Ensure the references are pertinent to the topic you are studying: {topic}. If there are no relevant references, output <references>[]</references>.

Now please output strictly in the following format:

\begin{minted}[breaklines, breakanywhere=true, breaksymbolleft={},
    breaksymbolright={}]{python}
<entities>{{A list of entities you extract}}</entities>
<idea>{{Background: ... \nNovelty: ...\nContribution:...\nMethods:...\nDetail reason:...\nLimitation:...\n }}</idea>
<experiment>{{Step1:... Step2:...}}</experiment>
<references>["{{Title1}}", "{{Title2}}",  ...]</references>
\end{minted}

\end{tcolorbox}

\clearpage
\section{Prompt for split the motivation and Experimental plan.}

\label{appendix:split_motivation_experiment}
\begin{tcolorbox}[breakable,enhanced,width=\textwidth]

\textbf{System Prompt:}  

You are required to act as an AI annotator and extract the methods mentioned or implied in the motivation and Experimental plan sections of the provided academic paper. The extraction should be done sentence by sentence, identifying any potential methods or techniques, whether explicitly stated or indirectly suggested.

A "method" or "technique" refers to a specific approach, algorithm, procedure, tool, or strategy mentioned or implied in the text. For each method, extract the core **keywords** that best represent the technique or approach. If a specific name of a method is mentioned, replace it with relevant keywords related to the technique. If the method is implied but not directly named, infer the core **keywords** associated with the method based on the context.

For each sentence, follow these guidelines:
\begin{itemize}
\item Identify the method(s) or technique(s) mentioned in the sentence.
\item If a specific name is used for a method, replace it with **keywords** related to the technique (such as “neural network,” “classification,” “transfer learning,” etc.).
\item If the method is implied but not directly named, infer the **keywords** that best describe the core functionality or application of the method.
\item Each method should be represented by a set of **keywords** that capture the essence of the technique, without unnecessary details.
\item Ensure that the **keywords** are distinct and representable in a simple form.
\end{itemize}
\vspace{0.3em}  
\textbf{User Instructions:}  
Please extract the methods mentioned or implied in the motivation and Experimental plan sections of the following academic paper. For each sentence, identify potential methods or techniques, whether explicitly stated or indirectly suggested.

Your response should be in JSON format, structured as follows:
\begin{minted}[breaklines, breakanywhere=true, breaksymbolleft={},
    breaksymbolright={}]{python}
{
    "motivation": [
        {
            "sentence": "<The sentence from the motivation section>",
            "methods": [
                "<Keyword 1>",
                "<Keyword 2>",
                ...
            ]
        },
        ...
    ],
    "experiment_plan": [
        {
            "sentence": "<The sentence from the Experimental plan section>",
            "methods": [
                "<Keyword 1>",
                "<Keyword 2>",
                ...
            ]
        },
        ...
    ]
}
\end{minted}

Input Example:
\begin{minted}[breaklines, breakanywhere=true, breaksymbolleft={},
    breaksymbolright={}]{python}
"State-of-the-art computer vision systems are trained to predict a fixed set of predetermined object categories. To handle new categories, transfer learning techniques are employed to adapt the model to new data."
\end{minted}

Your Answer Example (in JSON format):
\begin{minted}[breaklines, breakanywhere=true, breaksymbolleft={},
    breaksymbolright={}]{python}
{
    "motivation": [
        {
            "sentence": "State-of-the-art computer vision systems are trained to predict a fixed set of predetermined object categories.",
            "methods": [
                "training",
                "object classification",
                "feature extraction"
            ]
        }
    ],
    "experiment_plan": [
        {
            "sentence": "To handle new categories, transfer learning techniques are employed to adapt the model to new data.",
            "methods": [
                "transfer learning",
                "model adaptation",
                "domain adaptation"
            ]
        }
    ]
}
\end{minted}

Here is the motivation and Experimental plan:

[motivation and Experimental plan]
\end{tcolorbox}

\clearpage
\section{Prompt for motivation to motivation matching.}
\label{appendix:motivation_to_motivation}
\begin{tcolorbox}[breakable,enhanced,width=\textwidth]
\textbf{System Prompt:} 

You are an AI trained to rigorously evaluate the similarity between two research motivations by analyzing their structural alignment, theoretical foundations, and problem focus. Your task is to compare each pair of motivations and determine how closely they address the same core issues, challenges, or research gaps. Prioritize depth of analysis over superficial keyword matching.

\vspace{0.3em}  
\textbf{User Instructions:} 
For each pair of motivations provided:  
\begin{itemize}[itemsep=2pt,topsep=0pt,parsep=0pt]
\item \textbf{Analyze Core Issues}: Identify the central problem(s), challenge(s), or knowledge gap(s) each motivation emphasizes.  
\item \textbf{Compare Contextual Alignment}: Assess whether the motivations operate within the same domain, theoretical framework, or practical context.  
\item \textbf{Evaluate Structural/Theoretical Overlap}: Determine if they share methodologies, hypotheses, or foundational theories (even if applied differently).
\end{itemize}

Rating Criteria:  

Rate similarity on a scale of 1–5: 
\begin{enumerate}[itemsep=2pt,topsep=0pt,parsep=0pt]
\item \textbf{No Similarity}: Entirely distinct problems, contexts, and methodologies.  
\item \textbf{Low Similarity}: Minimal overlap in one aspect (e.g., tangential problem mention but divergent focus/theory).  
\item \textbf{Moderate Similarity}: Shared problem domain but differing approaches/theories (e.g., both address climate change mitigation but focus on policy vs. technology).  
\item \textbf{High Similarity}: Aligned problem focus and theory with minor differences in scope or application (e.g., both study AI bias in healthcare, but one targets diagnostics and the other patient data).  
\item \textbf{Complete Similarity}: Identical problems, theories, and contextual applications.  
\end{enumerate}

Output Format:

Return a JSON object for each pair with: 
\begin{itemize}[itemsep=2pt,topsep=0pt,parsep=0pt] 
\item A numeric `rating` (1–5)  
\item A concise `explanation` highlighting specific overlaps and key distinctions (1–2 sentences).
\end{itemize}

Example Output:
\begin{minted}[breaklines, breakanywhere=true, breaksymbolleft={},
    breaksymbolright={}]{python}
{
    "motivation_similarity": {
    "rating": 4,
    "explanation": "Both motivations address algorithmic bias in healthcare AI, but Motivation A focuses on diagnostic inaccuracies in radiology, while Motivation B emphasizes biases in patient prioritization systems."
    }
}
\end{minted}

Here are the motivations:

Motivation 1: [motivation1]

Motivation 2: [motivation2]

\end{tcolorbox}

\clearpage
\section{Prompt for Experimental plan to experiment matching.}
\label{appendix:experiment_to_experiment}
\begin{tcolorbox}[breakable,enhanced,width=\textwidth]
\textbf{System Prompt:}

You are an AI trained to evaluate the similarity between Experimental plans based on their structure, theory, and problem focus. Your task is to compare each pair of Experimental plan provided, focusing on the underlying issues and proposed solutions. Ignore the order of steps or naming conventions used in the plans. Instead, assess if the plans tackle similar problems or use comparable theoretical frameworks.

\vspace{0.3em}  
\textbf{User Instructions:}

Compare the two provided Experimental plans and assess their similarity using the criteria below.

Evaluation Criteria: 
\begin{itemize}[itemsep=2pt,topsep=0pt,parsep=0pt]
\item \textbf{Structural Similarity:}
\begin{itemize}
   \item Do the plans share a comparable framework (e.g., hypothesis testing, control groups, data analysis methods)?  
   \item Ignore differences in step order or terminology.  
\end{itemize}
\item \textbf{Theoretical Alignment:} Do they rely on the same or related theories, principles, or methodologies?  
\item \textbf{Problem Focus:} Do they address the same underlying problem or challenge, even if solutions differ?  
\end{itemize}

Rating Scale (1–5): 
\begin{enumerate}[itemsep=2pt,topsep=0pt,parsep=0pt]
\item \textbf{No similarity:} Entirely different problems/theories/structures.  
\item \textbf{Low similarity:} Minor overlap in one criterion (e.g., same problem but divergent theories).  
\item \textbf{Moderate similarity:} Align in 1–2 criteria with clear differences (e.g., shared theory but distinct structures).  
\item \textbf{High similarity:} Align in 2–3 criteria with minor discrepancies (e.g., same problem and structure but different theories).  
\item \textbf{Complete similarity:} Identical in all criteria.  
\end{enumerate}
Output Requirements:  
\begin{itemize}
\item  Return JSON format with `rating` (1–5) and `explanation`.  
\item  The explanation must explicitly reference \textbf{structure}, \textbf{theory}, and \textbf{problem focus}. 
\end{itemize}

Example Output:  

\begin{minted}[breaklines, breakanywhere=true, breaksymbolleft={},
    breaksymbolright={}]{python}
{
  "experiment_plan_similarity": {
    "rating": 3,
    "explanation": "Both address energy efficiency (problem focus) and use quantitative metrics (structure), but Plan 1 employs machine learning (theory) while Plan 2 uses statistical modeling (theory)."
  }
}
\end{minted}

Here are the Experimental plans:

Experimental plan 1: [experiment\_plan1]

Experimental plan 2: [experiment\_plan2]
\end{tcolorbox}

\clearpage
\section{Prompt for idea to topic matching.}
\label{appendix:idea_topic}
\begin{tcolorbox}[breakable,enhanced,width=\textwidth]
\textbf{System Prompt:} 

You are a highly advanced AI tasked with evaluating research papers. When asked to evaluate the alignment between the motivation and Experimental plan with the topic of the paper, follow these guidelines:

\begin{itemize}
\item Motivation: Check if the rationale clearly explains why this topic is being explored and how it connects to the broader context of the field. Identify the research gap or problem being addressed and ensure that it directly connects with the topic.
\item Experimental plan: Assess whether the experiment design is directly aligned with the motivation. Determine if the methods proposed are appropriate for addressing the research questions identified in the motivation and if they suit the topic under investigation.
\end{itemize}

After evaluating, assign a compatibility score (1-5) based on the following scoring rubric:

\vspace{0.3em}  
\textbf{User Instructions:}  
Given the research paper's topic, evaluate whether the motivation and Experimental plan are aligned with the focus of the study. Provide a compatibility score from 1 to 5, where:
\begin{enumerate}
\item (Very Poor Alignment): The motivation and/or Experimental plan are completely unrelated to the topic. The motivation does not address the topic, and the Experimental plan does not effectively test the research question posed.
\item (Poor Alignment): The motivation and/or Experimental plan are weakly connected to the topic. The motivation addresses the topic but in a vague or unclear manner, and the Experimental plan partially, but not effectively, tests the research question.
\item (Moderate Alignment): The motivation and Experimental plan are somewhat aligned with the topic. The motivation explains the topic, but with some gaps in clarity, and the Experimental plan mostly addresses the research question, though some methods may not be optimal.
\item (Good Alignment): The motivation and Experimental plan are clearly aligned with the topic. The motivation explains the research problem well, and the Experimental plan is suitable and addresses the research question effectively.
\item (Excellent Alignment): The motivation and Experimental plan are perfectly aligned with the topic. The motivation provides a strong, clear rationale for the study, and the Experimental plan is highly suitable and directly addresses the research question in a methodologically sound way.
\end{enumerate}

Please output your evaluation in JSON format with the following structure:

Output Format:

Return a JSON object for each pair with:  
- A numeric `rating` (1–5)  
- A concise `explanation` highlighting specific overlaps and key distinctions (1–2 sentences).

\begin{minted}[breaklines, breakanywhere=true, breaksymbolleft={},
    breaksymbolright={}]{python}
{{"motivation": {{
    "alignment": <score (1-5)>,
    "comments": "<concise evaluation of motivation's alignment with topic>"
  }},
  "experiment_plan": {{
    "alignment": <score (1-5)>,
    "comments": "<concise evaluation of Experimental plan's alignment with topic>"
  }}}}
\end{minted}

Ensure the output contains the specific evaluations for the motivation, Experimental plan, and an overall compatibility score.

Here are the input:
Topic: [topic] Motivation: [motivation] Experimental plan: [experiment\_plan]
\end{tcolorbox}

\clearpage
\section{Prompt for MCQ motivation.}
\label{appendix:MCQ_motivation}
\begin{tcolorbox}[breakable,enhanced,width=\textwidth]
\textbf{System Prompt:} 

You are an AI motivation analyzer. Your task is to compare a user-provided motivation against four options (A, B, C, D) and identify the closest match. Follow these steps:
\begin{itemize}
\item \textbf{Analyze the Input:} Extract the core problem, theme, or goal from the input motivation.  
\item \textbf{Compare Options:} Evaluate each option (A-D) by identifying its primary focus (problem addressed, underlying theme, or end goal).  
\item \textbf{Match Criteria:} Prioritize matches based on shared intent, not just keywords. Discard superficial similarities.  
\item \textbf{Explain Clearly:} Highlight specific overlapping elements (e.g., "both target efficiency improvement" or "address ethical concerns").  
\end{itemize}
\vspace{0.3em}  
\textbf{User Instructions:} 

Return a JSON response with: 
\begin{itemize}
\item \textbf{closest\_option:} The best-matched option (A-D).  
\item \textbf{explanation:} A concise, evidence-based comparison (1-2 sentences).  
\end{itemize}

Input Motivation: [input\_motivation]
\begin{minted}[breaklines, breakanywhere=true, breaksymbolleft={},
    breaksymbolright={}]{python}
Options:
- A: [option_a]
- B: [option_b]
- C: [option_c]
- D: [option_d]
\end{minted}
Provide your response in JSON format as follows:
\begin{minted}[breaklines, breakanywhere=true, breaksymbolleft={},
    breaksymbolright={}]{python}
{  
"closest_option": "A",  
"explanation": "{input_motivation} aligns with Option A because both [specific shared problem/theme/goal]. For example, [concrete similarity]."  
}
\end{minted}
\end{tcolorbox}

\clearpage
\section{Prompt for MCQ Experimental plan.}
\label{appendix:MCQ_experiment_plan}
\begin{tcolorbox}[breakable,enhanced,width=\textwidth]
\textbf{System Prompt:} 

You are an expert in comparative analysis of scientific methodologies. Your task is to match an input Experimental plan to the most similar of four predefined plans (A, B, C, D) by evaluating theoretical alignment and problem-solving focus.

Analysis Guidelines:
\begin{itemize}
\item \textbf{Ignore:} 
   \begin{itemize}
   \item Step order and step names.  
   \item Superficial differences in terminology.  
   \end{itemize}
\item \textbf{Compare using these criteria:} 
\begin{itemize}
   \item \textbf{Structural \& Theoretical Alignment:} Core framework, methodology organization, and foundational principles.    
   \item  \textbf{Problem Focus:} Type of challenges addressed, objectives prioritized, and domain-specific issues targeted.  
\end{itemize} 
\item \textbf{Output Requirements:} \begin{itemize}
   \item \textbf{Return JSON with two keys:} `closest\_plan` (A-D) and `explanation` (3-5 sentences).    
   \item  In the explanation, explicitly address both structural/theoretical alignment **and** problem focus. 
\end{itemize}
 
\end{itemize}
\vspace{0.3em}  
\textbf{User Instructions:}

Analyze this input Experimental plan:  
[input\_plan]

\begin{minted}[breaklines, breakanywhere=true, breaksymbolleft={},
    breaksymbolright={}]{python}
Compare it against these predefined plans:
- A: {plan_a}  
- B: {plan_b}  
- C: {plan_c}  
- D: {plan_d}  
\end{minted}
Provide your response in JSON format as follows:

\begin{minted}[breaklines, breakanywhere=true, breaksymbolleft={},
    breaksymbolright={}]{python}
{
"closest_plan": "C",
"explanation": "The input plan {input_plan} aligns with Option C structurally through... Simultaneously, both address... [Always mention BOTH criteria]"
}
\end{minted}

\end{tcolorbox}

\clearpage
\section{Prompt for idea competition.}
\label{appendix:MCQ_idea_competition}
\begin{tcolorbox}[breakable,enhanced,width=\textwidth]
\textbf{System Prompt:} 

You are a judge in a competition. Your task is to evaluate and compare two ideas based on their motivation and Experimental plans. You must decide which idea is better according to the evaluation criteria provided. Be objective, avoid biases, and ensure your decision is based solely on the quality of the ideas and experiments.

\vspace{0.3em}  
\textbf{User Instructions:} 

The motivation and Experimental plan of idea0 is: [motivation0, experiment\_plan0]

The motivation and Experimental plan of idea1 is: [motivation1, experiment\_plan1]

The topic of the competition is: [topic]

Which motivation and experiment do you think is better? Please write a short paragraph to explain your choice.

Here are your evaluation criteria:

1. Novelty: Are the problems or approaches new? Is this a novel combination of familiar techniques? Is it clear how this work differs from previous contributions? Is related work adequately referenced?

2. Significance: Are the ideas important? Are other people (practitioners or researchers) likely to use these ideas or build on them? Does the idea address a difficult problem in a better way than previous research? Does it provide a unique theoretical or pragmatic approach?

3. Quality: Is there a clear rationale for each step of the experimental design? Are the baseline and evaluation metrics chosen appropriately? Has the design taken into account the potential advantages and limitations of the methods used? Can this experimental design effectively support the claims made in the idea.

4. Feasibility: Can the idea be realized with existing technology or methods? Are there any technical difficulties or bottlenecks? Is the idea clear and logical? Is there any obvious error or unreasonable part in the idea, and can the experiment be designed normally according to this idea. 

5. Clarity: Is the motivation and experiment clearly written? Does it provide enough information for the expert reader to understand the experiment? Is it well organized? Does it adequately inform the reader?

6. Relevance to Topic: Does the idea align with the provided topic? Does it address the core themes or issues raised by the topic? How closely is the idea tied to the overall focus of the competition?

Note: Avoid any position biases and ensure that the order in which the responses were presented does not influence your decision. DO NOT allow the LENGTH of the responses to influence your evaluation, choose the one that is straight-to-the-point instead of unnecessarily verbose. Be as objective as possible. (very important!!!)

If you think idea0 is better than idea1, you should output 0. If you think idea1 is better than idea0, you should output 1.

Your output should be strictly in the following JSON format:
\begin{minted}[breaklines, breakanywhere=true, breaksymbolleft={},
    breaksymbolright={}]{python}
{{
    "Novelty_choice": {{
        "thinking_process": "Your detailed reasoning here...",
        "choice": <Your choice (0 or 1)>
    }},
    "Significance_choice": {{
        "thinking_process": "Your detailed reasoning here...",
        "choice": <Your choice (0 or 1)>
    }},
    "Feasibility_choice": {{
        "thinking_process": "Your detailed reasoning here...",
        "choice": <Your choice (0 or 1)>
    }},
    "Clarity_choice": {{
        "thinking_process": "Your detailed reasoning here...",
        "choice": <Your choice (0 or 1)>
    }},
    "Relevance_choice": {{
        "thinking_process": "Your detailed reasoning here...",
        "choice": <Your choice (0 or 1)>
    }}

    "Final_choice": <Your choice (0 or 1)>
}}
\end{minted}

\end{tcolorbox}

\clearpage

\end{document}